\documentclass[runningheads]{llncs}
\usepackage{graphicx}
\usepackage{tikz}
\usepackage{comment}
\usepackage{amsmath,amssymb} % define this before the line numbering.
\usepackage{color}

\usepackage[accsupp]{axessibility}  % Improves PDF readability for those with disabilities.

\usepackage{graphicx}
\usepackage{amsmath}
\usepackage{amssymb}
\usepackage{booktabs}

\usepackage{bbm}
\usepackage[ruled,linesnumbered]{algorithm2e}
\usepackage{setspace}

\usepackage{tabularx}
\usepackage{booktabs}
\usepackage{multirow}
\usepackage{diagbox}
\usepackage{nicematrix}
\usepackage[pagebackref,breaklinks,colorlinks]{hyperref}

\makeatletter
\@namedef{ver@everyshi.sty}{}
\makeatother
\usepackage{tikz}
\usetikzlibrary{spy}

\tikzset{
    cross/.pic = {
    \draw (-#1,0) -- (#1,0);
    \draw (0,-#1) -- (0,#1);
    }
}

\usepackage{xcolor}
\usepackage{color}
\usepackage{colortbl}

\def\eg{\emph{e.g.}} 
\def\ie{\emph{i.e.}}

\def\etal{\emph{et al.}}

\makeatletter
\newcommand{\rmnum}[1]{\romannumeral #1}
\newcommand{\Rmnum}[1]{\expandafter\@slowromancap\romannumeral #1@}
\makeatother

\begin{document}
% \linenumbers
\pagestyle{headings}
\mainmatter
\def\ECCVSubNumber{1759}  % Insert your submission number here

\title{MPIB: An MPI-Based Bokeh Rendering Framework for Realistic Partial Occlusion Effects} % Replace with your title

\begin{comment}
\titlerunning{ECCV-22 submission ID \ECCVSubNumber} 
\authorrunning{ECCV-22 submission ID \ECCVSubNumber} 
\author{Anonymous ECCV submission}
\institute{Paper ID \ECCVSubNumber}
\end{comment}
%******************

% \begin{comment}
\titlerunning{MPIB: An MPI-Based Bokeh Rendering Framework}
\author{Juewen Peng\inst{1} \and Jianming Zhang\inst{2} \and Xianrui Luo\inst{1} \and Hao Lu\inst{1} \and Ke Xian\inst{1}\thanks{Corresponding author} \and Zhiguo Cao\inst{1}}
\authorrunning{J. Peng et al.}
\institute{Key Laboratory of Image Processing and Intelligent Control, Ministry of Education,\\School of AIA, Huazhong University of Science and Technology, China \\
\email{\{juewenpeng,zgcao,xianruiluo,hlu,kexian\}@hust.edu.cn}
\and Adobe Research \\
\email{jianmzha@adobe.com} \\
\url{https://github.com/JuewenPeng/MPIB}
} 
%******************
\maketitle

\begin{abstract}
Partial occlusion effects are a phenomenon that blurry objects near a camera are semi-transparent, resulting in partial appearance of occluded background. However, it is challenging for existing bokeh rendering methods to simulate realistic partial occlusion effects due to the missing information of the occluded area in an all-in-focus image. Inspired by the learnable 3D scene representation, Multiplane Image (MPI), we attempt to address the partial occlusion by introducing a novel MPI-based high-resolution bokeh rendering framework, termed MPIB. To this end, we first present an analysis on how to apply the MPI representation to bokeh rendering. Based on this analysis, we propose an MPI representation module combined with a background inpainting module to implement high-resolution scene representation. This representation can then be reused to render various bokeh effects according to the controlling parameters. To train and test our model, we also design a ray-tracing-based bokeh generator for data generation. Extensive experiments on synthesized and real-world images validate the effectiveness and flexibility of this framework.
\keywords{Bokeh Rendering, Multiplane Image, Partial Occlusion}
\end{abstract}

\begin{figure}[t]
    \setlength{\abovecaptionskip}{5pt}
    \small
	\centering
	\renewcommand\arraystretch{1.2}
    \begin{tabular}{*{4}{c@{\hspace{.7mm}}}}
        \includegraphics[width=0.24\linewidth]{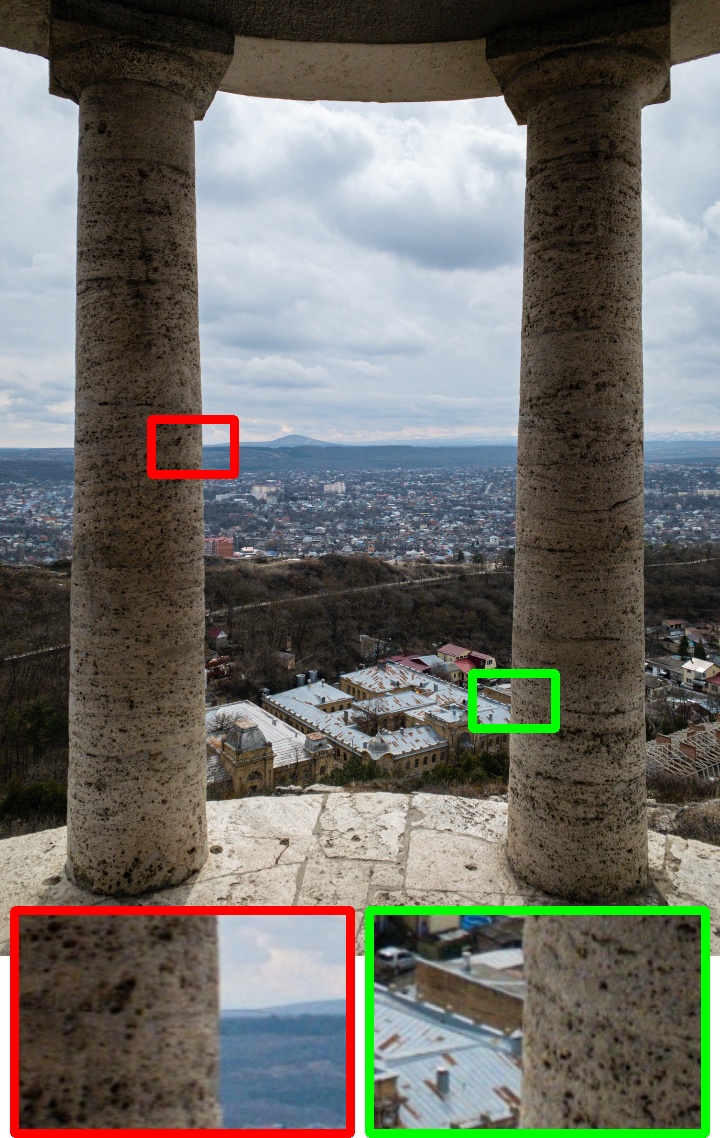} &
       \includegraphics[width=0.24\linewidth]{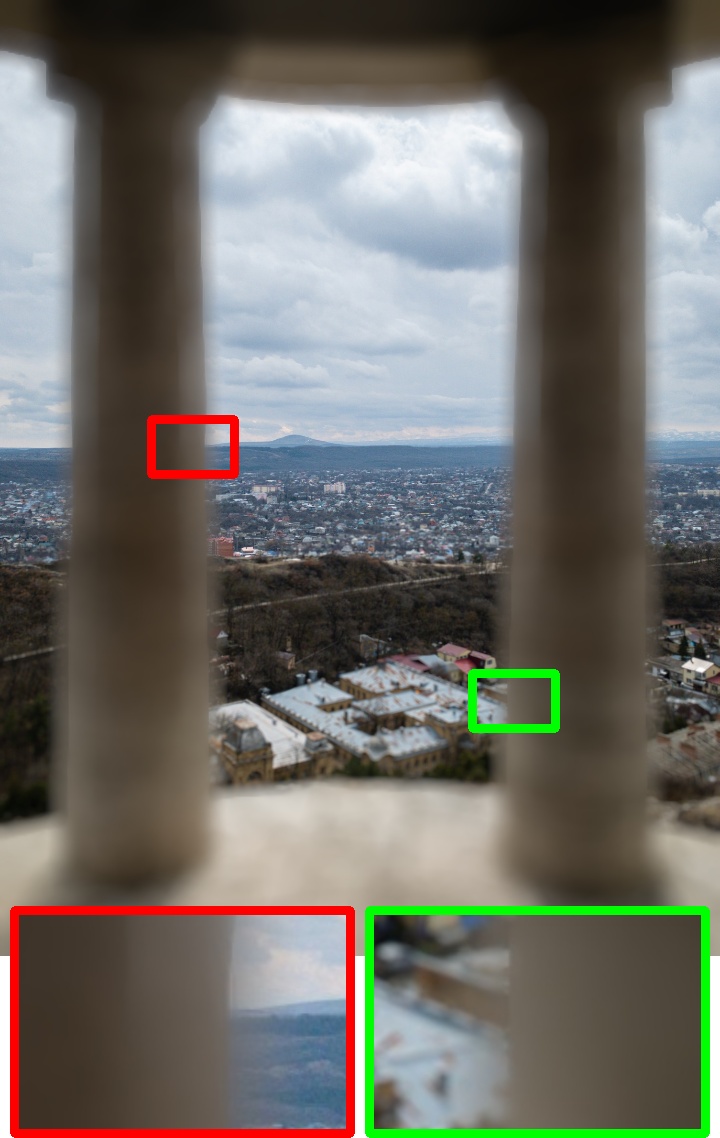} &
        \includegraphics[width=0.24\linewidth]{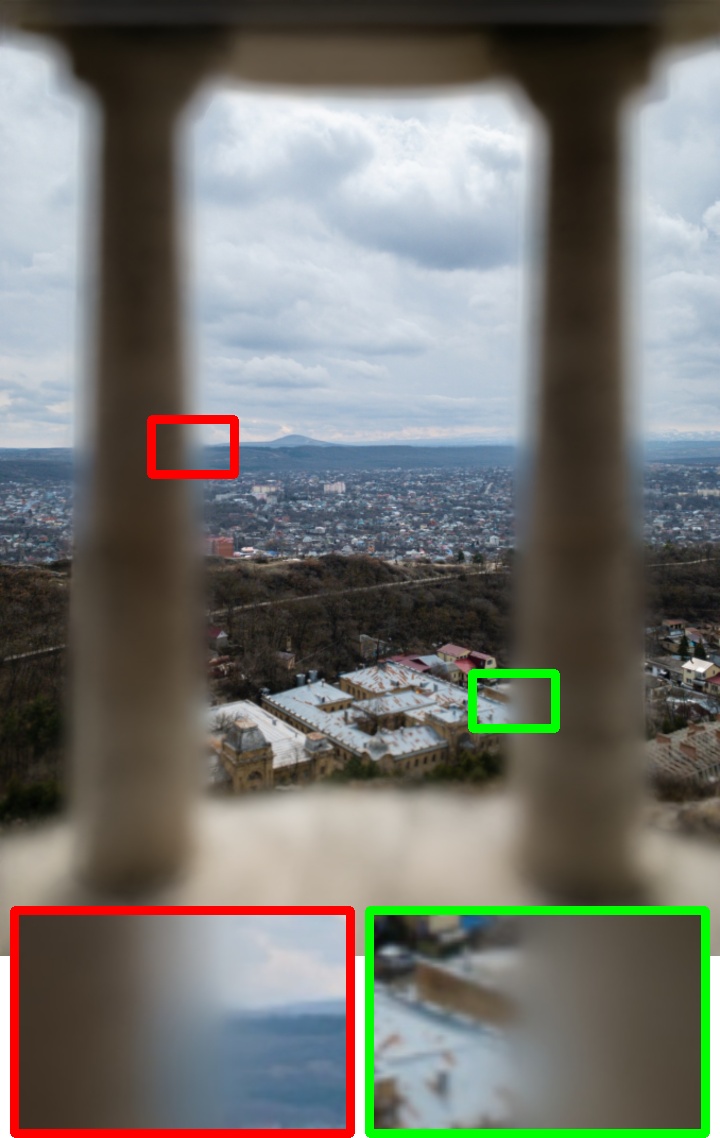} &
        \includegraphics[width=0.24\linewidth]{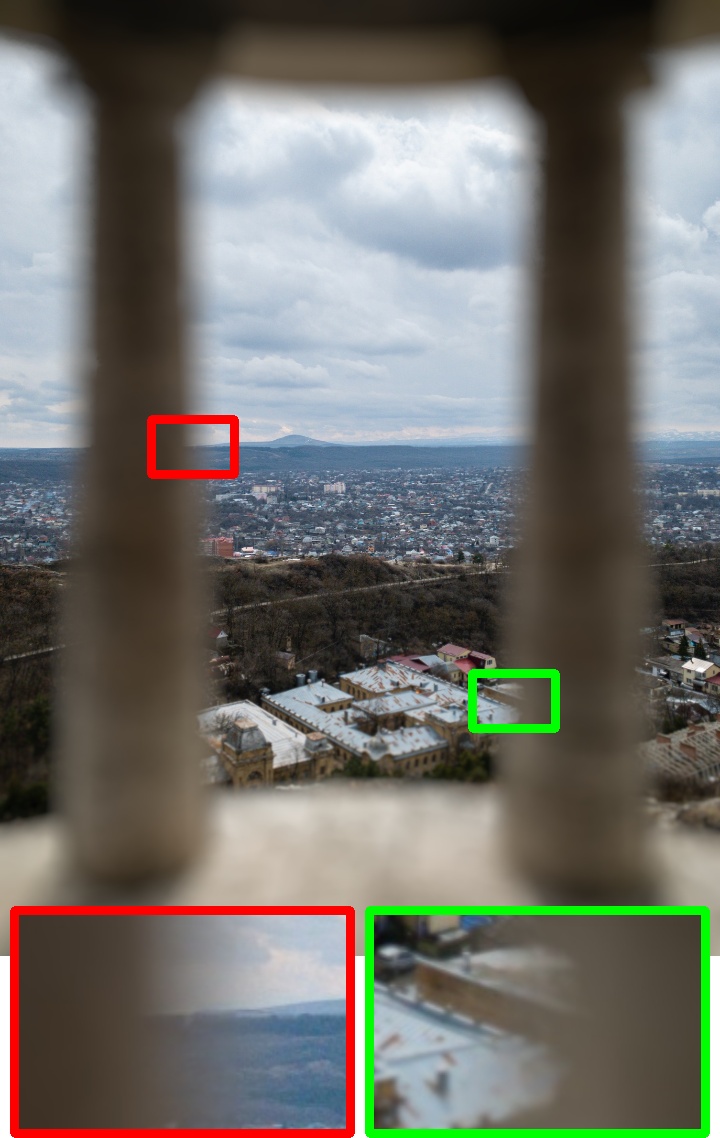} \\
        
        \includegraphics[width=0.24\linewidth]{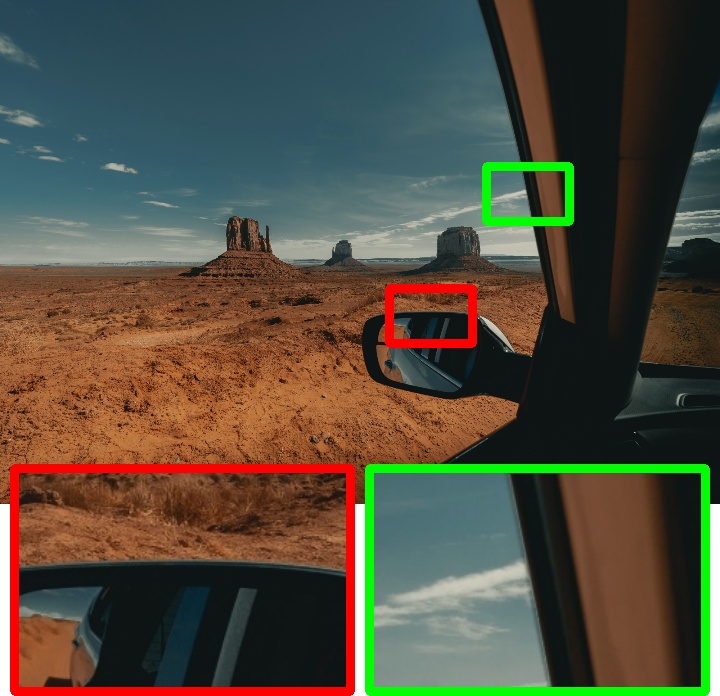} &
        \includegraphics[width=0.24\linewidth]{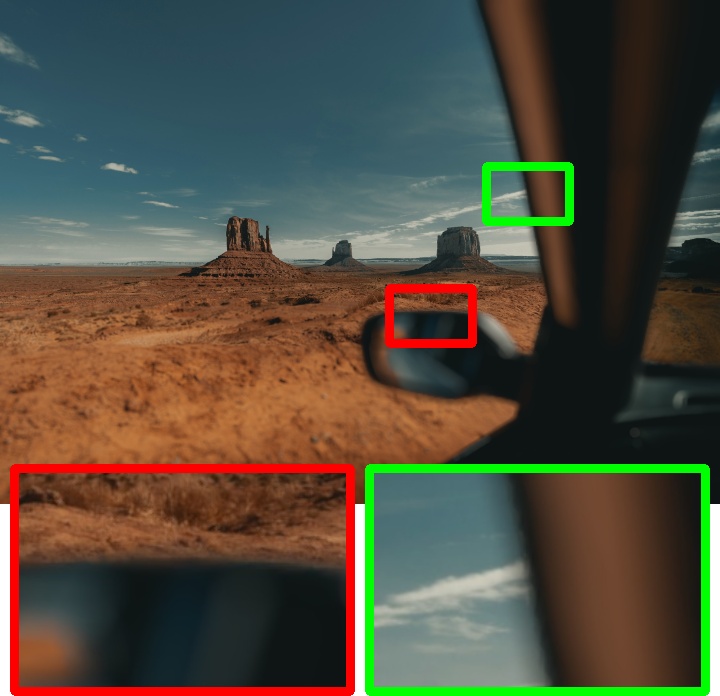} &
        \includegraphics[width=0.24\linewidth]{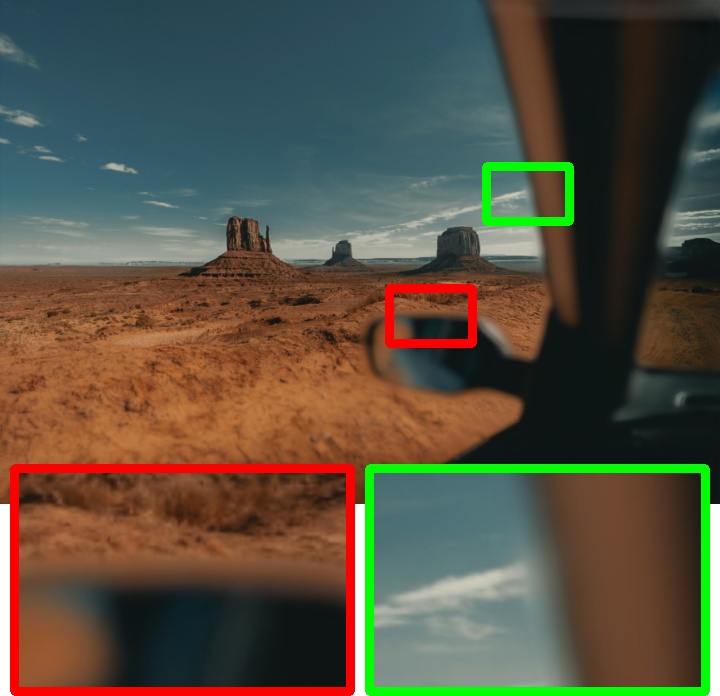} &
        \includegraphics[width=0.24\linewidth]{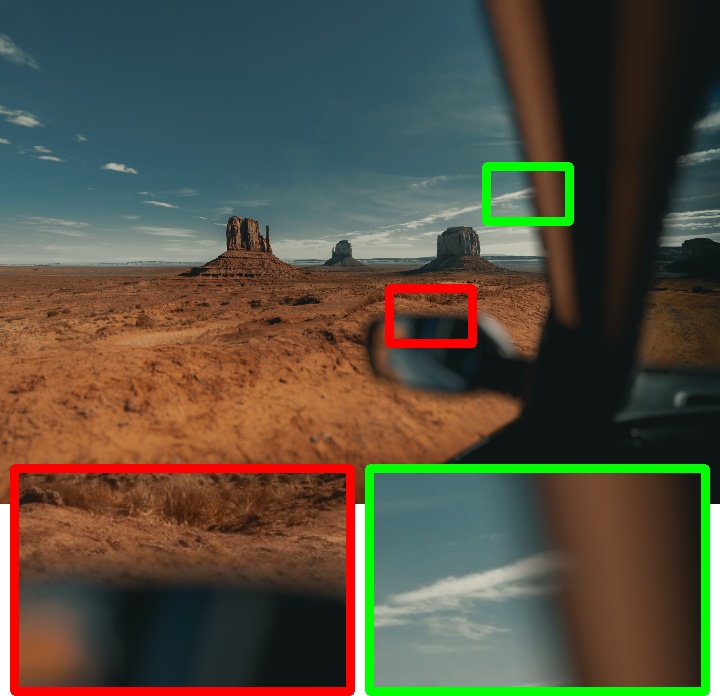} \\
 
        All-in-Focus & SteReFo~\cite{busam2019sterefo} & DeepLens~\cite{wang2018deeplens} & MPIB (Ours) \\
    \end{tabular}
    \caption{MPIB renders more realistic partial occlusion effects than other methods at the boundary between foreground and background. Best viewed by zooming in.
    }
    \label{fig:teaser}
\end{figure}

\section{Introduction}
Bokeh effect is commonly used in photography to create an appealing blur effect in out-of-focus areas and make the subject stand out from the picture. Various post-processing methods, which simulate the bokeh effect from an all-in-focus image, have been proposed so far. However, as shown in Fig.~\ref{fig:teaser}, neither the physically based ones (\eg, SteReFo~\cite{busam2019sterefo}) nor the deep learning-based ones (\eg, DeepLens~\cite{wang2018deeplens}) can well handle the issue of partial occlusion: the out-of-focus object close to the camera becomes blurred and semi-transparent, revealing the occluded in-focus area.

In the field of computer graphics, Schedl and Wimmer~\cite{schedl2012layered} first attempt to solve this problem by decomposing the scene into different depth layers, and blurring each layer with a fixed-size filter before compositing them together. This strategy works well given the complete scene information, however, it is less effective for methods with a single image as input~\cite{busam2019sterefo,zhang2019synthetic}. A recent idea relevant to this layering strategy is Multiplane Image (MPI) proposed for novel view synthesis~\cite{zhou2018stereo}. MPI aims to learn a 3D scene representation with multiple RGBA planes from a single image, which can then be used to synthesize different novel views of the scene. Despite its desirable characteristics of explicitly modeling occluded areas for each plane, we observe that current MPI-based view synthesis methods~\cite{li2021mine,srinivasan2019pushing,tucker2020single,zhou2018stereo} are still inadequate in restoring the occluded surface with complicated textures.

To apply MPI to bokeh rendering and address the above challenges, we first analyse the difference of MPI representation and layer compositing formulation between the view synthesis and bokeh rendering. Then, based on this analysis, we propose a novel MPI-based framework, termed MPIB, for synthetic bokeh effect. Specifically, we combine an MPI representation module with a background inpainting module to obtain a high-resolution 3D scene representation. The background inpainting module aims to synthesize convincing contents in the occluded areas and lighten the burden of the scene representation. Once the scene representation is obtained, it can be reused to render different bokeh effects with adjustable controlling parameters, such as blur amount and refocused disparity.

Due to the difficulty of capturing accurate pairs of all-in-focus image and bokeh image in the real world, we design a ray-tracing-based bokeh generator for training. We also use it to synthesize a test dataset for the initial validation. Since the judgement of the bokeh effect is really subjective, we further conduct a user study on images collected from websites and compare our results with the latest iPhone 13 Cinematic Mode. Experimental results show that MPIB renders realistic partial occlusion effects and yields substantial improvements over state-of-the-art methods.

In summary, our main contributions are as follows.
\begin{itemize}
    \item[$\bullet$] We present an analysis on how to apply the MPI representation and the layer composting scheme to bokeh rendering.
    \item[$\bullet$] We propose an MPI-based framework for high-resolution bokeh rendering and realistic partial occlusion effects.
    \item[$\bullet$] We design a ray-tracing-based bokeh generator, which creates almost real bokeh effects and can be used to produce training and test data.
\end{itemize}

\section{Related Work}
% \noindent\textbf -> subsubsection
\subsubsection{Bokeh Rendering.} Bokeh rendering techniques can be classified into physically based methods and neural rendering methods. 
In the early years, most physically based methods~\cite{abadie2018advances,lee2010real,schedl2012layered,wu2013rendering,yu2010real} entail 3D scene information and are time-consuming. Recent methods~\cite{barron2015fast,busam2019sterefo,hach2015cinematic,peng2021interactive,shen2016automatic,shen2016deep,wadhwa2018synthetic,xian2021ranking,yang2016virtual,zhang2019synthetic}, which render bokeh effects from a single image and a corresponding depth map, are more efficient and practical. One classic idea is layered rendering~\cite{busam2019sterefo,zhang2019synthetic}, \ie, decomposing the scene into multiple layers and independently blurring each layer before compositing them from back to front. However, they do not consider potential depth inaccuracy and object occlusion, resulting in unrealistic partial occlusion effects.

To solve these problems, many neural rendering methods have been proposed recently. Xiao \etal~\cite{xiao2018deepfocus} specialize in using a perfect depth map to render realistic bokeh effects in low resolution. Considering the difficulty of obtaining perfect depth maps in the real world, Wang \etal~\cite{wang2018deeplens} propose a robust rendering system consisting of the depth prediction, lens blur, and guided upsampling modules. To further simplify the rendering process, many end-to-end networks~\cite{dutta2021stacked,ignatov2019aim,ignatov2020aim,ignatov2020rendering,qian2020bggan} are proposed. They simulate the bokeh effect of DSLR cameras from a single wide depth-of-field image without inputting the depth map or any other controlling parameters and show a compelling performance on the bokeh dataset EBB!~\cite{ignatov2020rendering}. However, the simplicity comes at a cost. These networks are lack of flexibility. They cannot adjust bokeh effects, such as different blur amounts and focal planes. In this work, we focus on controllable bokeh rendering with an all-in-focus image, a potentially imperfect disparity map and some controlling parameters as input.

\subsubsection{MPI Representation.} Since Zhou \etal~\cite{zhou2018stereo} first propose MPI to reconstruct the camera frustum from stereo images and synthesize novel views, this representation has been widely used in novel view synthesis methods~\cite{li2021mine,mildenhall2019local,srinivasan2019pushing,tucker2020single,zhou2016view} due to the appealing property of differentiability and explicitly modeling occluded contents. Srinivasan \etal~\cite{srinivasan2019pushing} provide a theoretical analysis of MPI limits, and use the appearance flow~\cite{zhou2016view} to improve the realism in occluded areas. Tucker~\etal~\cite{tucker2020single} apply the MPI representation to the single-view inputting case which is more practical but more challenging. Li~\etal~\cite{li2021mine} propose an encoder-decoder architecture which is a continuous depth generalization of MPI by introducing the idea of neural radiance fields (NeRF)~\cite{mildenhall2020nerf}. Different from the above works, we embed an additional disparity map into the model aside from the all-in-focus image and apply a lightweight guided upsampling network. It aims to enhance the generalization of the model and obtain the high-resolution scene representation with accurate and sharp object boundaries. In addition, we supervise our model with bokeh images instead of synthesized images in different views.

\subsubsection{Image Inpainting.} The goal of inpainting is to fill in missing contents of an image. Compared with traditional patch-based~\cite{criminisi2003object} and nearest neighbor-based~\cite{hays2007scene} methods, neural networks have much stronger ability to capture spatial context of images and generate plausible contents for unseen parts, espcially after the introduction of adversarial training~\cite{goodfellow2014generative}. Therefore, deep learning-based inpainting has attracted a lot of attention and many CNN-based methods have been proposed~\cite{iizuka2017globally,pathak2016context,yang2017high}. Recent works attempt to replace the regular convolutions with partial~\cite{liu2018image}, gated~\cite{yu2019free} or fourier~\cite{suvorov2022resolution} convolutions, and design new architectures~\cite{liu2020rethinking,nazeri2019edgeconnect,zhu2021image} to address irregular masks and high-resolution inpainting. In this work, we utilize the off-the-shelf inpainting model~\cite{suvorov2022resolution} to facilitate our MPI representation and create more convincing partial occlusion effects.

\section{MPIB: An MPI-Based Bokeh Rendering Framework}
\begin{figure}[t]
\setlength{\abovecaptionskip}{-5pt}
\begin{center}
\includegraphics[width=\linewidth]{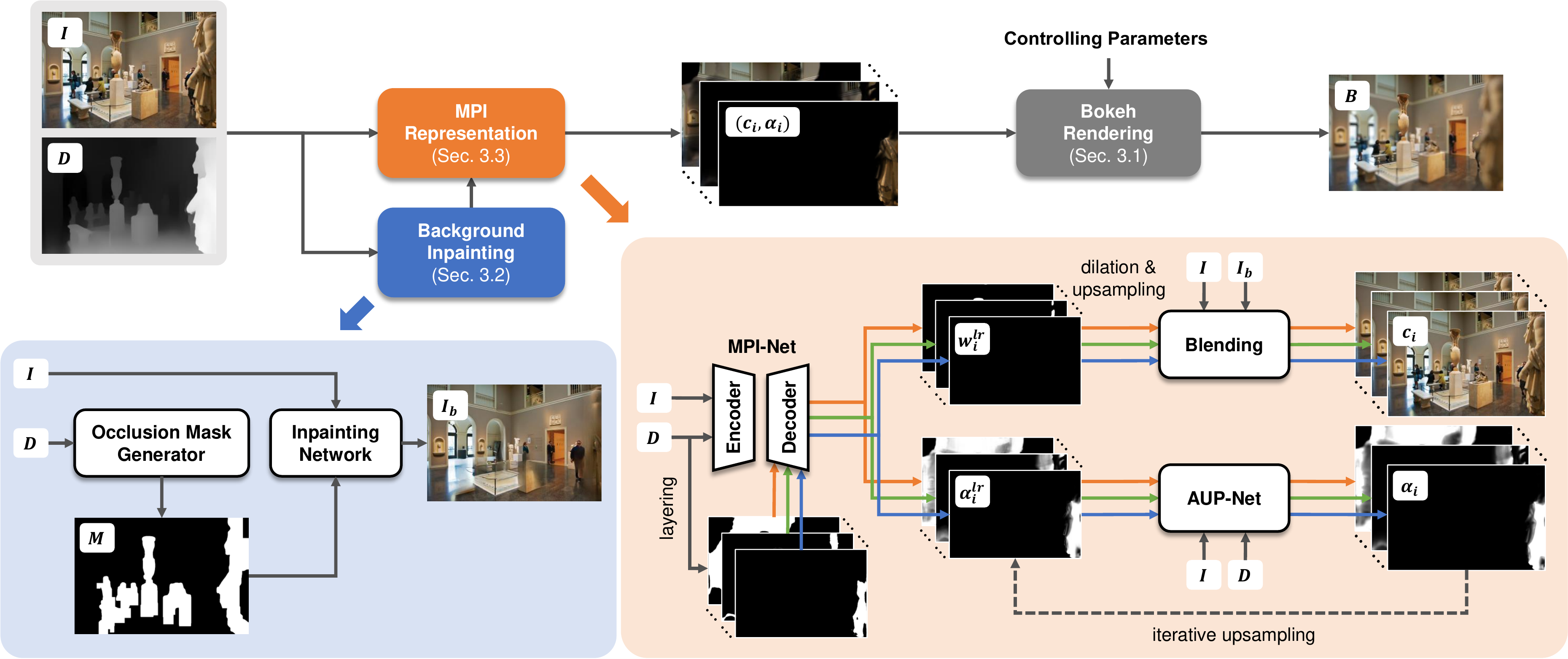}
\end{center}
\caption{
Our framework MPIB takes an all-in-focus image and a potentially imperfect disparity map as input to obtain a 3D scene representation of multiple RGBA planes. This representation is generated by an MPI representation module and a background inpainting module. Then, the scene representation can be reused to produce multiple bokeh images according to different controlling parameters.}
\label{fig:framework}
\end{figure}

As shown in Fig.~\ref{fig:framework}, our framework consists of $3$ modules: background inpainting, MPI representation, and bokeh rendering. In the following, we first analyse how to apply MPI in bokeh rendering and introduce our rendering formula (Sec.~\ref{sec:bokeh_rendering}). Then, we describe the structures of our background inpainting module (Sec.~\ref{sec:background_inpainting}) and MPI representation module (Sec.~\ref{sec:mpi_representation}). Finally, we provide the training details, including the proposed ray-tracing-based bokeh generator (Sec.~\ref{sec:model_training}).

\subsection{MPI Representation and Layer Compositing Formulation}
\label{sec:bokeh_rendering}
The well-known multiplane image (MPI) representation, introduced by Zhou \etal~\cite{zhou2018stereo}, consists of a set of fronto-parallel planes. Each plane encodes an RGB color image $c_i$ and
an alpha map $\alpha_i$, \ie, $\{(c_i,\alpha_i)\,|\,i=1,2,...,N\}$, where $N$ is the number of MPI planes. When applying the MPI representation to bokeh rendering, we make minor modifications. As proven in~\cite{wadhwa2018synthetic,yang2016virtual}, the blur radius $r$ of each pixel is
\begin{equation}\label{eq:radius}
    r=A\,\Big\lvert \frac{1}{z}-\frac{1}{z_f} \Big\rvert=A\,\lvert d-d_f \rvert\,,
\end{equation}
where $A$ reflects the overall blur amount of the image. $z$ is the depth of the pixel and $z_f$ is the refocused depth. We replace the depth $z$ with the disparity (inverse depth) $d$ to simplify the formula. One can see that $r$ is the linear variation of $d$. For uniform sampling of the blur amount across different planes, we replace depth discretization in general MPI representation with disparity discretization.

In novel view synthesis, one can reconstruct the scene $I$ by continuously using the “over” alpha compositing operation~\cite{levoy1990efficient,porter1984compositing} to MPI planes with a back-to-front order:
\begin{equation}\label{eq:image_from_mpi}
    I=\sum^N_{i=1}\Big(c_i\alpha_i\prod^N_{j=i+1}(1-\alpha_j)\Big)\,.
\end{equation}
When applying Eq.~\ref{eq:image_from_mpi} to the bokeh rendering, we blur each layer with a fixed-size kernel $K_i$,  which is adaptive to the controlling parameters. The rendered bokeh image $B$ can be formulated as
\begin{equation}\label{eq:bokeh_from_mpi}
    B =\sum^N_{i=1}\Big((c_i\alpha_i*K_i)\prod^N_{j=i+1}(1-\alpha_j*K_j)\Big)\,,
\end{equation}
where $*$ is the convolution operation. To eliminate the artifacts caused by discretization, we add extra weight normalization to Eq.~\ref{eq:bokeh_from_mpi} as follows:
\begin{equation}\label{eq:bokeh_from_mpi_norm}
    B=\frac{\sum^N_{i=1}\Big((c_i\alpha_i*K_i)\prod^N_{j=i+1}(1-\alpha_j*K_j)\Big)}{\sum^N_{i=1}\Big((\alpha_i*K_i)\prod^N_{j=i+1}(1-\alpha_j*K_j)\Big)}\,.
\end{equation}

Typically, the transformation from scene irradiance to RGB values is nonlinear~\cite{yang2016virtual}, so we conduct the gamma transformation to $c_i$ and the inverse gamma transformation to $B$ following common practice.
% A common practice is to apply the gamma correction before and after the rendering. To this end, we conduct the gamma transformation to $c_i$ and the inverse transformation to $B$.

\begin{figure}[t]
    \setlength{\abovecaptionskip}{5pt}
    \scriptsize
	\centering
	\renewcommand\arraystretch{1.2}
	\begin{tabular}{*{6}{c@{\hspace{.7mm}}}}
        \includegraphics[width=0.16\linewidth]{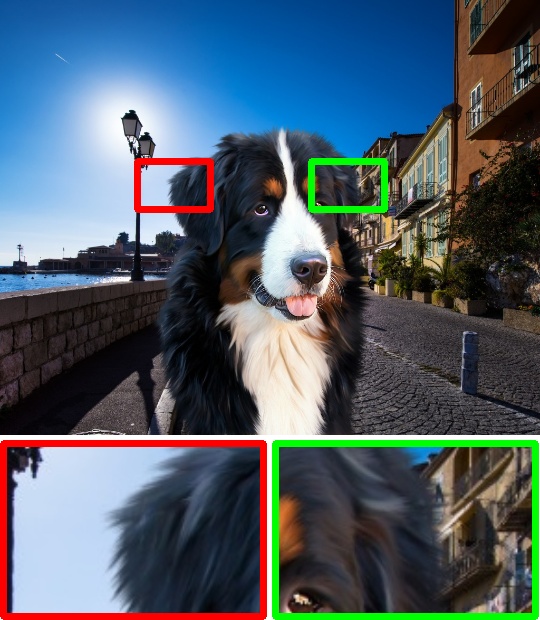} &
        \includegraphics[width=0.16\linewidth]{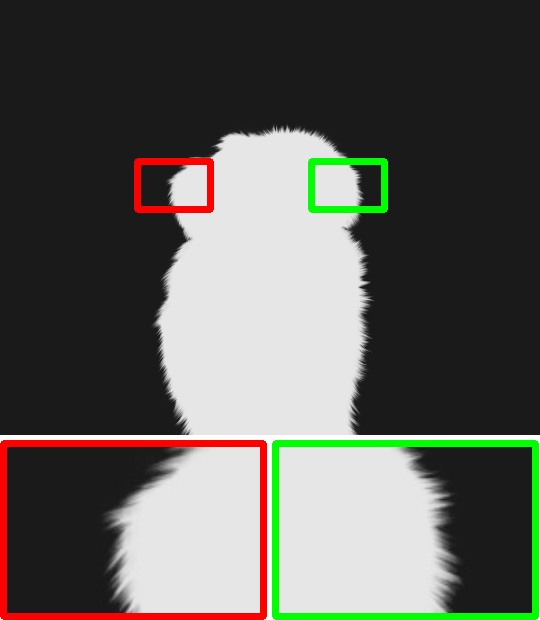} &
        \includegraphics[width=0.16\linewidth]{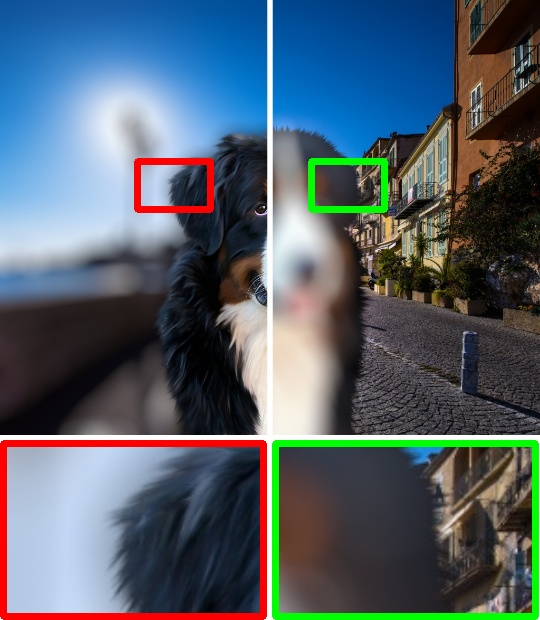} &
        \includegraphics[width=0.16\linewidth]{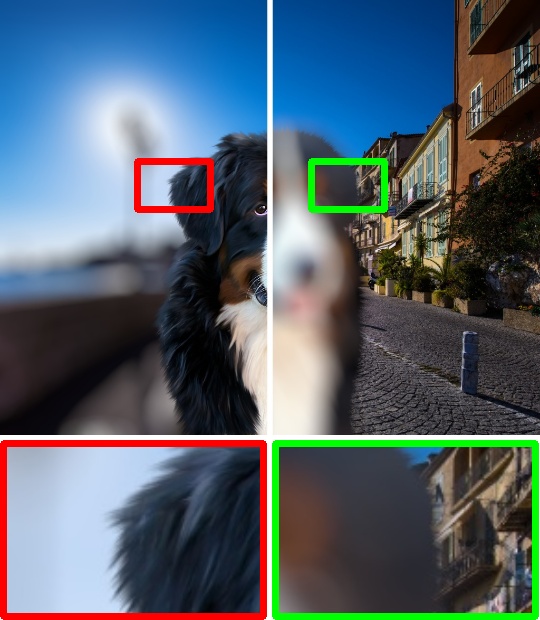} &
        \includegraphics[width=0.16\linewidth]{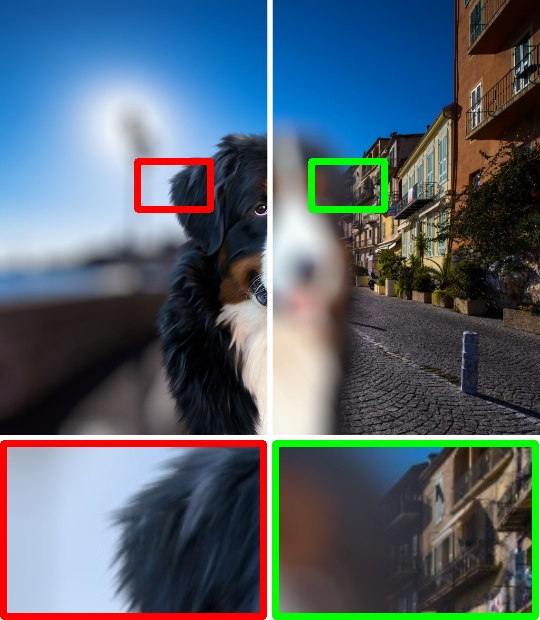} &
        \includegraphics[width=0.16\linewidth]{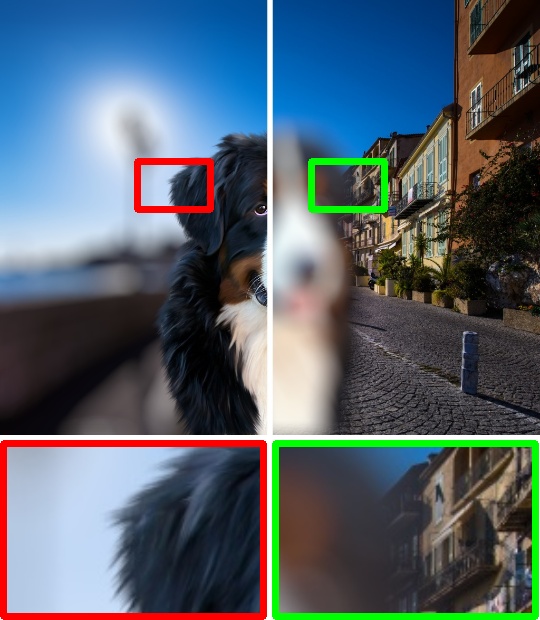} \\
        
        All-in-Focus & Disparity & RVR~\cite{zhang2019synthetic} & SteReFo~\cite{busam2019sterefo} & Ours & GT \\
	\end{tabular}
	\caption{Toy experiment. The scene is synthesized from two images with constant disparities. ``GT'' refers to the result of the ray-tracing-based bokeh generator proposed in Sec.~\ref{sec:model_training}. For each rendered bokeh image, the left half is refocused on the foreground while the right half is refocused on the background.}
	\label{fig:toy}
\end{figure}

Previous layered rendering methods, such as RVR~\cite{zhang2019synthetic} and SteReFo~\cite{busam2019sterefo}, adopt a similar compositing strategy. However, they only consider visible parts of the image and calculate the alpha map of each plane manually. 
% We formulate their rendering pipelines in the supplementary material. 
For clarity, we conduct a toy experiment in Fig.~\ref{fig:toy} where only two planes are used. We compare our approach with RVR, SteReFo, and the bokeh generator proposed in Sec.~\ref{sec:model_training} (regarded as the ground truth). Note that the result of our approach is in an ideal situation with known background information. One can see that our result is most similar to the ground truth, and the occluded background is partly visible at the boundary of the blurred foreground, demonstrating the plausibility of our rendering formula and the importance of predicting occluded contents.

\subsection{Background Inpainting Module}
\label{sec:background_inpainting}
In background inpainting module, we combine an off-the-shelf inpainting model LaMa~\cite{suvorov2022resolution} with an occlusion mask generator to produce a background image. Since occlusion often occurs where the disparity of the scene changes significantly and the occluded area is on the side with a larger disparity, we design the occluded mask generator according to this principle.

\begin{figure}[t]
    \setlength{\abovecaptionskip}{5pt}
    \scriptsize
	\centering
	\renewcommand\arraystretch{1.2}
	\begin{tabular}{*{6}{c@{\hspace{.5mm}}}}
        \includegraphics[width=0.16\linewidth]{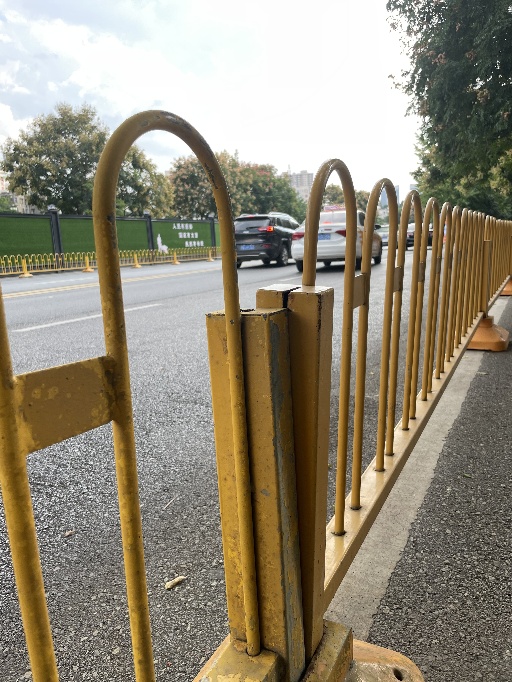} &
        \includegraphics[width=0.16\linewidth]{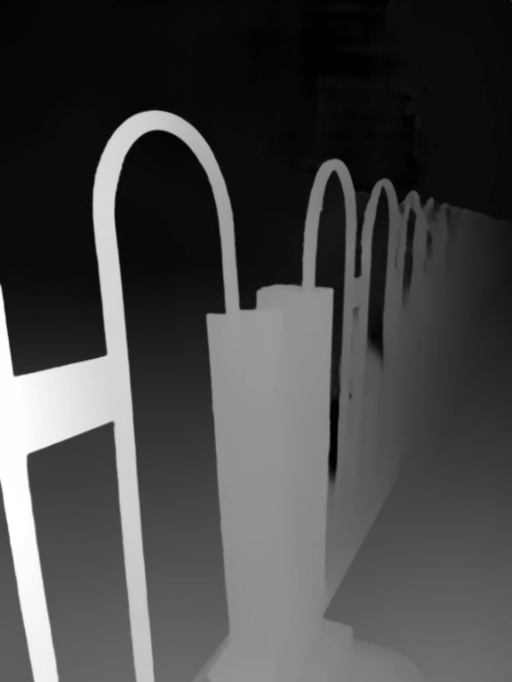} &
        \includegraphics[width=0.16\linewidth]{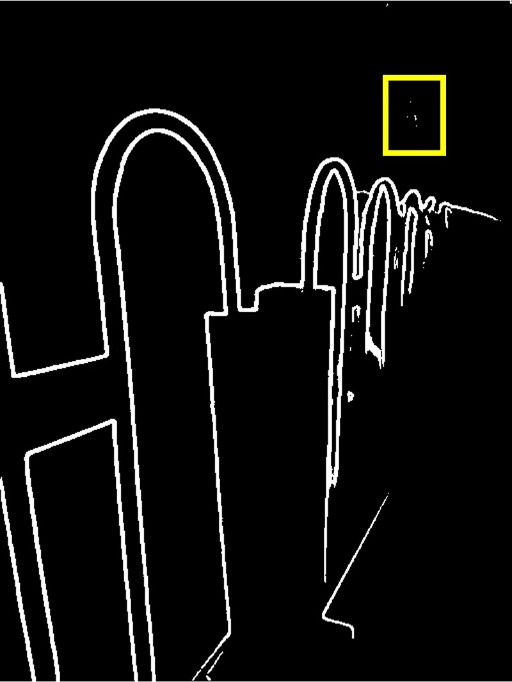} &
        \includegraphics[width=0.16\linewidth]{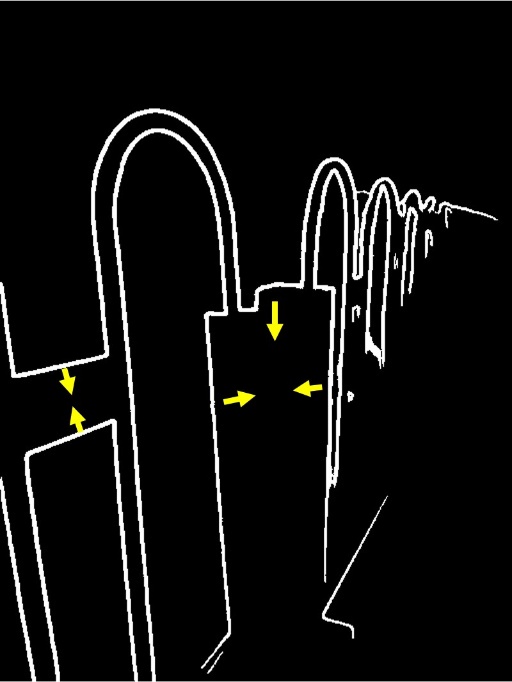} &
        \includegraphics[width=0.16\linewidth]{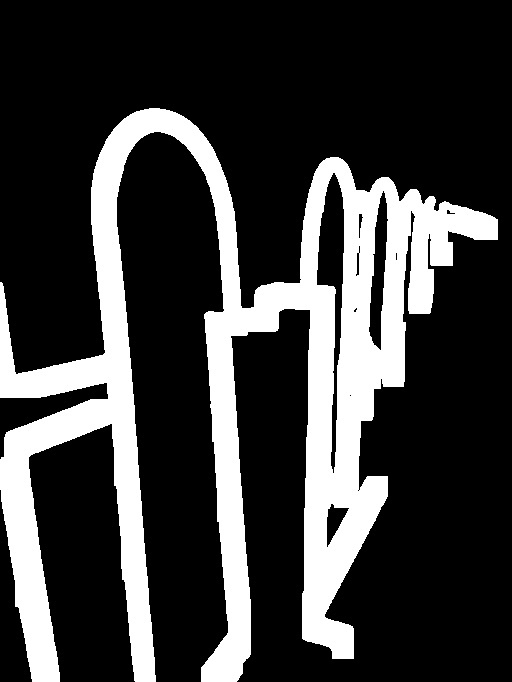} &
        \includegraphics[width=0.16\linewidth]{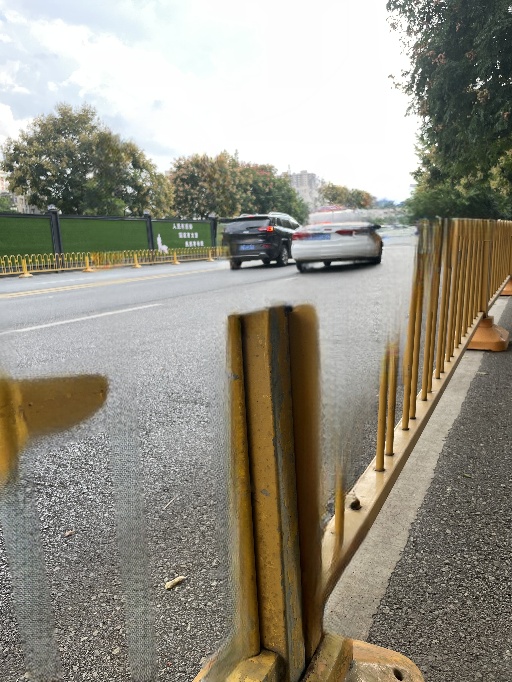} \\
        All-in-Focus & Disparity & ${\rm Mask_1}$ & ${\rm Mask_2}$ & ${\rm Mask_3}$ & Inpainted BG \\
	\end{tabular}
	\caption{Generation of the occlusion mask and the inpainted background image. ``${\rm Mask_1}$'' is calculated by thresholding the gradient of the disparity map. ``${\rm Mask_2}$'' removes the short segments of ``${\rm Mask_1}$''. ``${\rm Mask_3}$'' extends the area of ``${\rm Mask_2}$'' in the direction of disparity increase. ``Inpainted BG'' is predicted by the inpainting method LaMa~\cite{suvorov2022resolution} using the all-in-focus image and the produced occlusion mask as input.}
	\label{fig:occlusion_mask}
\end{figure}

At the beginning, we use the Sobel operator to calculate a $2$-channel gradient map $G=\{G_x,G_y\}$ of the input disparity map $D$. By thresholding the gradient magnitude, we obtain a depth discontinuity mask and regard it as the initial occlusion mask $M$. Then, we remove short segments of $M$ and $G$ to reduce the impact of noise. Subsequently, we iteratively extend $M$ in the direction of gradient increase. In each iteration, we first perform $\ell_2$ normalization to $G$ to obtain unit normal vectors $G^n$. Then, we forward warp $G^n$ via the softmax splatting operation~\cite{niklaus2020softmax} using the same $G^n$ as the optical flow to get the inner ring of $G^n$, and represent it with $G^{w}$. At last, we update $G$ by
\begin{equation}\label{eq:update_grad}
    G = M \cdot G^n + (1-M) \cdot G^{w}\,,
\end{equation}
and update $M$ to refer to the nonzero areas of $G$. In practice, as the input disparity map may not align well with the all-in-focus image at occluding boundaries, we finally dilate $M$ by several pixels as processed in~\cite{shih20203d} to prevent the foreground color leakage and reduce inpainting artifacts. 
% The complete algorithm is in the supplementary material. 
For ease of understanding, we visualize the process of producing an occlusion mask in Fig.~\ref{fig:occlusion_mask}. 

Note that for complicated scenes, we can restrict the area of occlusion mask and predict more background images for MPI planes in different disparity levels. However, we show in Sec.~\ref{sec:real_world} that one background image is sufficient for rendering real-world images in general.

% \begin{algorithm}[!t]
%     % \footnotesize
%     \small
%     % \setstretch{1.1}
%     \caption{Occlusion Mask Generator}
%     \label{alg:occlusion_mask}
    
%     \SetKwFunction{Sobel}{Sobel}\SetKwFunction{RemoveShortSegments}{RemoveShortSegments}\SetKwFunction{Sample}{Sample}
    
%     \KwIn{Disparity map $D$}
%     \KwOut{Occlusion mask $M$}
%     % \setlength{\baselineskip}{10pt}
    
%     Calculate the gradient $G$ of disparity map $D$ by the Sobel operator\,\;
%     Generate the initial occlusion mask $M$ by thresholding the magnitude of $G$\,\;
%     Remove short segments of $M$ and $G$\,\;
%     \For {$i\leftarrow 1$ \KwTo $T$}
%     {
%         $\ell_2$ normalization to $G$ to obtain $G^n$\,\;
%         Forward warp $G^n$ by softmax splatting operation~\cite{niklaus2020softmax} using the same $G^n$ as the optical flow to obtain $G^{w}$\,\;
%         Update $G$ by $G \leftarrow M \cdot G + (1-M) \cdot G^w$\,\;
%         Update $M$ by thresholding the magnitude of $G$\,\;
%     }
%     Slightly dilate $M$ to obtain the final output\,\;

%     % $G \leftarrow$ \Sobel{$D$}\,\;
%     % $M \leftarrow \mathbbm{1}(G>\zeta)$\,\;
%     % $M \leftarrow$ \RemoveShortSegments{$M$}\,\;
%     % $G \leftarrow M\cdot G$\,\;
% \end{algorithm}

\subsection{High-resolution MPI Representation Module}
\label{sec:mpi_representation}
In MPI representation module, we first propose an encoder-decoder network MPI-Net to obtain an initial MPI representation as with~\cite{li2021mine}. For each image, the encoder runs only once, while the decoder runs multiple times to generate $N$ planes. However, there are two main differences to be aware of. (\rmnum{1}) We embed the disparity map $D$ into the decoder instead of the discrete disparity values to pass more prior knowledge to the network, which will lead to the 
better layering and stronger generalization. Specifically, assume $D$ is ranged from $0$ to $1$, we divide the range by $\{[\frac{i-1}{N},\frac{i}{N}]\,|\,i=1,2,...,N\}$, and calculate the coarse zone mask for each plane. Subsequently, we apply a single convolution layer to each zone mask to increase channels before feeding them into the decoder. (\rmnum{2}) To reduce the difficulty of training MPI-Net, we assume that the RGB image $c_i$ of each plane is the per-pixel weighted average of the original all-in-focus image $I$ and the inpainted background image $I^b$. Thus, for each plane, MPI-Net only predicts an alpha map $\alpha_i$ and a blend weight map $w_i$, and $c_i$ can be produced by
\begin{equation}
    c_i=w_i\cdot I^b+(1-w_i)\cdot I\,.
\end{equation}

To adapt our model to high-resolution input, we also propose a lightweight guided upsampling network AUP-Net to iteratively upsample the alpha map $\alpha_i$ by a factor of $2$. Note that AUP-Net is guided by the high-resolution all-in-focus image $I$ and its corresponding disparity map $D$. As for the blend weight map $w_i$, we just perform slight dilation and bilinear upsampling. The reason for the dilation here is similar to the reason we dilated the occlusion mask earlier in Sec.~\ref{sec:background_inpainting}. Although this operation may cause a small amount of visible information loss on $c_i$, it effectively prevents foreground colors from bleeding into background planes and reduces the risk of unpleasant boundary artifacts.

\begin{figure}[t]
\setlength{\abovecaptionskip}{-5pt}
\setlength{\belowcaptionskip}{-5pt}
\begin{center}
\includegraphics[width=0.85\linewidth]{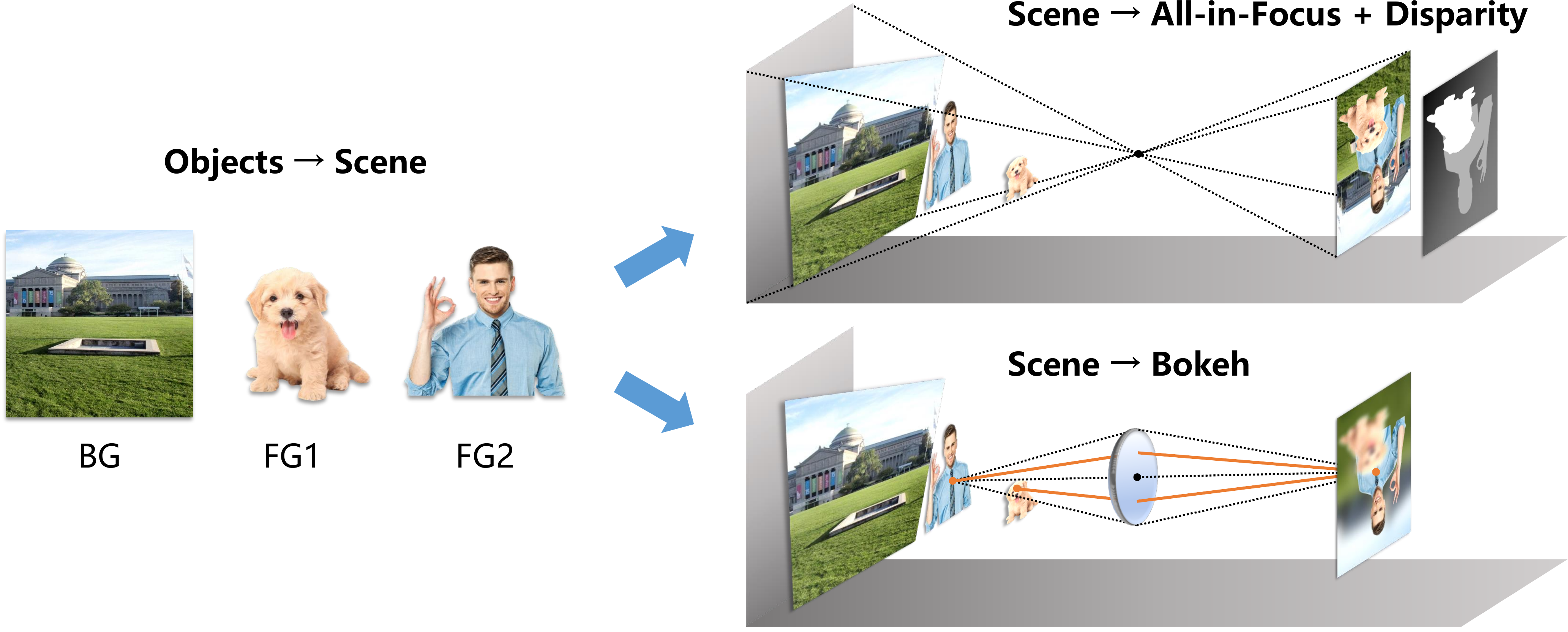}
\end{center}
\caption{
Pipeline of ray-tracing-based bokeh generator. Given a background RGB image and two foreground RGBA images, we first construct a 3D scene through randomly setting the positions and disparities of different objects. Then, we can synthesize an all-in-focus image and a corresponding disparity map by compositing different object planes, and generate a bokeh image by backward tracing the path of light rays.}
\label{fig:bokeh_generator}
\end{figure}

\subsection{Model Training}
\label{sec:model_training}
\subsubsection{Bokeh Generator and Training Data.} To obtain ground-truth bokeh images, we design a ray-tracing-based bokeh generator (Fig.~\ref{fig:bokeh_generator}), which is capable of creating a bokeh image from a background RGB image and some foreground RGBA images with customized disparities and different controlling parameters. Specifically, for each image to be composited, we set its disparity map $d$ as a plane equation of pixel coordinates $(x, y)$:
\begin{equation}
    d = \frac{1-ax-by}{c}\,,
\end{equation}
where $a$, $b$, $c$ are three constants. We project all images to a 3D scene based on their disparities, and treat them as different objects. For each pixel $(x, y)$ of the rendered result, we sample $2500$ rays passing through the aperture, look for the intersection of each ray and the scene, and project the intersection to the sensor plane. The detailed calculation is in the supplementary material. Here, we directly provide the coordinates $(x_n, y_n)$ of the projected intersection:
\begin{equation}\label{eq:ray_tracing}
    \left\{
    \begin{aligned}
        x_n & = x + \frac{1 - ax - by - cd_f}{a A\mu + b A\nu + c}\,A\mu\,,\\
        y_n & = y + \frac{1 - ax - by - cd_f}{a A\mu + b A\nu + c}\,A\nu\,,
    \end{aligned}
    \right.
\end{equation}
where $A$ is the blur parameter. $d_f$ is the refocused disparity. $(\mu, \nu)$ denotes the sampling point within the aperture, which meets $\mu^2+\nu^2\leq1$. $(x_n, y_n)$ is available if and only if it is inside the object. Specially, if $(x_n, y_n)$ happens to be on the boundary, we suppose the ray with a proportion equal to the alpha value of $(x_n, y_n)$ intersects the current plane, and the ray with the remaining proportion continues to propagate. In summary, we calculate the intersection of the ray and each plane from front to back, and the search process stops until the energy of the ray is exhausted. The final rendered result of $(x, y)$ is the weighted average of the colors for all intersections.

In practice, we first randomly select $20$k images from Places~\cite{zhou2017places} as our background images. The foreground images originate from two sources. One is PhotoMatte85~\cite{lin2021real} with $85$ portrait RGBA images. The other is websites. We collect $300$ images with pure background and extract their alpha maps by matting tools. During the training, each sample is synthesized online from $2$ random foreground images and $1$ background image with the resolution of $256\times256$. The disparity map is set within the range from $0$ to $1$. The ground-truth bokeh image is produced with the random blur parameter from $0$ to $32$, refocused disparity from $0$ to $1$, and gamma value from $1$ to $4$. To improve the generalization of the model, we augment the input disparity map with random noise, gaussian blur, dilation and erosion. We also use the intermediate synthesized results instead of the inpainted results as $I^b$ to simplify and accelerate the training process.

\subsubsection{Loss Functions.} The training of MPI-Net and AUP-Net are independent. For MPI-Net, we use the following loss:
\begin{equation}
    \mathcal{L}_{\mathit{MPI}} = \mathcal{L}_{bokeh} + 0.4\,\mathcal{L}_{disp} + 0.1\,\mathcal{L}_{alpha} + 0.1\,\mathcal{L}_{weight}\,,
\end{equation}
where $\mathcal{L}_{bokeh}$ is a $\ell_1$ loss, which encourages the rendered bokeh image $B$ to match the ground truth $B^*$ in both image space and gradient space. $\mathcal{L}_{disp}$ has the same form with $\mathcal{L}_{bokeh}$, which enforces the reconstructed disparity map $D^{rc}$ to be consistent with the raw input disparity map $D^*$ without augmentation. This term aims to assist MPI representation learning and improve the sensitivity of the model to object boundaries. As with~\cite{tucker2020single}, the reconstructed disparity is defined by
\begin{equation}\label{eq:re_disp}
    D^{rc}=\sum^N_{i=1}\Big(d_i\alpha_i\prod^N_{j=i+1}(1-\alpha_j)\Big)\,,
\end{equation}
where $d_i=\frac{i-0.5}{N}$, which represents the average disparity of each plane. Next, $\mathcal{L}_{alpha}$ and $\mathcal{L}_{weight}$ are two $\ell_1$ regularization losses, which constrain the predicted alpha maps and blend weight maps to be smooth, respectively.

For training AUP-Net, we use a similar loss:
\begin{equation}
    \mathcal{L}_{\mathit{AUP}} = \mathcal{L}_{bokeh} + 0.4\,\mathcal{L}_{disp} + 0.1\,\mathcal{L}_{alpha} + 0.4\,\mathcal{L}_{\mathit{self}}\,.
\end{equation}
$\mathcal{L}_{weight}$ is not used here because AUP-Net only processes alpha maps. Instead, we add a self-supervised loss $\mathcal{L}_{\mathit{self}}$ to accelerate the convergence and prevent the upsampled alpha map $\alpha_i$ and the input low-resolution alpha map $\alpha_i^{lr}$ from being too different.
\begin{equation}
    \mathcal{L}_{\mathit{self}} = \mathcal{L}_{\ell_1}\big({\rm downsample}(\alpha_i),\alpha_i^{lr}\big)\,.
\end{equation}

\subsubsection{Implementations.} We implement our model by PyTorch~\cite{paszke2017automatic}. The number of MPI planes is set to $32$. In MPI-Net, we use ResNet-18 as the encoder, and use a decoder similar to~\cite{li2021mine}. AUP-Net is a lightweight network based on U-Net~\cite{ronneberger2015u}. We show its detailed architecture in the supplementary material. When training AUP-Net, we forcely downsample the input by a factor of $2$ before applying MPI-Net, and use AUP-Net to upsample the initial rendered result up to the original resolution. Both networks are trained for $40$ epochs using the Adam optimizer~\cite{kingma2015adam} with a learning rate of $10^{-4}$, and a batch size of $8$. All experiments are conducted on 
four NVIDIA GeForce GTX 1080 Ti GPUs.

\section{Experiments}
\subsection{Bokeh Rendering on Synthesized Dataset}
\label{sec:syn_data}
\subsubsection{Dataset.} Current public bokeh datasets including EBB!~\cite{ignatov2020rendering}, Aperture~\cite{zhang2019synthetic} and a Unity synthesized dataset~\cite{xiao2018deepfocus} are unsuitable for evaluations of controllable bokeh rendering. The first two datasets are manually captured by a DSLR camera with unknown controlling parameters. There exist color inconsistency and scene misalignment between the wide and shallow DoF image pairs, and almost all images are focused on the front subject. For the last synthesized dataset, all scenes are built from randomized object geometries, so there is a huge gap between them and real-world images. In addition, the maximum blur size of this dataset is too small. As a result, we use our ray-tracing-based generator to synthesize a test dataset, which contains $100$ new scenes (not the same with our training data) with the resolution of $1024\times1024$. Each scene is synthesized from $1$ background image and $3$ foreground images. Unlike~\cite{xiao2018deepfocus}, all images are derived from the real world. We provide two versions for disparity settings. In the first version, the disparity of each object is set to a constant value, while in the second version, it is set to a smoothly varying plane. Besides, for each all-in-focus image, we synthesize $16$ bokeh images with $4$ blur parameters from $20$ to $80$ and $4$ refocused disparities, which correspond to the disparities of the $4$ objects. The gamma values are set to $2.2$ for all settings.

\subsubsection{Metrics.} To measure performance, we use LPIPS~\cite{zhang2018unreasonable}, PSNR and SSIM as metrics. Since the difficulty of bokeh rendering is mainly concentrated at occluding boundaries and the human eye is more sensitive to these areas, we additionally introduce two metrics ${\rm PSNR_{ob}}$ and ${\rm SSIM_{ob}}$, which only reflect the accuracy at occluding boundaries. More details are in the supplementary material.

\subsubsection{Compared Methods.} We compare MPIB with $4$ physically based rendering methods, including $2$ pixel-wise rendering methods: Gather~\cite{wadhwa2018synthetic} and Scatter~\cite{wadhwa2018synthetic}, and $2$ layered rendering methods: RVR~\cite{zhang2019synthetic} and SteReFo~\cite{busam2019sterefo}. We also test $2$ neural rendering methods: DeepLens~\cite{wang2018deeplens} and DeepFocus~\cite{xiao2018deepfocus}. Note that as DeepFocus~\cite{xiao2018deepfocus} cannot handle large blur sizes, directly applying it to high resolution will cause the collapse of the model. Thus, we downsample the input before feeding it into the network, and the rendered result is bilinearly upsampled without refining. More discussion about DeepFocus is in the supplementary material. To be fair, we discard the depth estimation modules in SteReFo~\cite{busam2019sterefo} and DeepLens~\cite{wang2018deeplens}, and only preserve their rendering modules. Then, all methods can take the same all-in-focus image, disparity map and controlling parameters as input.

\begin{table}[t]
    % \setlength{\abovecaptionskip}{5pt}
    % \setlength{\belowcaptionskip}{-10pt}
    % \small
	\centering
	\caption{Quantitative results on the synthesized dataset. $N_{bg}$ denotes the number of the inpainted images used in our background inpainting module. The best performance is in \textbf{boldface}, and the second best is \underline{underlined}.} 
	\resizebox{1.0\linewidth}{!}{
    \setlength{\tabcolsep}{4pt}
	\renewcommand\arraystretch{1.0}
	\begin{NiceTabular}{l|ccccc|ccccc}
		\toprule
		\multicolumn{1}{l}{\multirow{2}{*}[-0.5ex]{Method}} & \multicolumn{5}{c}{Constant disparity for each object} & \multicolumn{5}{c}{Varying disparity for each object} \\
		\cmidrule{2-11} % \cmidrule(r){2-6} \cmidrule(r){7-11} 
		~ & LPIPS$\downarrow$ & PSNR$\uparrow$ & ${\rm PSNR_{ob}}\!\!\uparrow$ & SSIM$\uparrow$ & ${\rm SSIM_{ob}}\!\!\uparrow$ & LPIPS$\downarrow$ & PSNR$\uparrow$ & ${\rm PSNR_{ob}}\!\!\uparrow$ & SSIM$\uparrow$ & ${\rm SSIM_{ob}}\!\!\uparrow$ \\
		\midrule
		\midrule
		Gather~\cite{wadhwa2018synthetic} & 0.042 & 32.6 & 25.4 & 0.978 & 0.896 & 0.044 & 33.0 & 25.9 & 0.979 & 0.903 \\
		Scatter~\cite{wadhwa2018synthetic} & 0.029 & 33.9 & 26.7 & 0.983 & 0.921 & 0.027 & 34.8 & 27.8 & 0.985 & 0.934 \\
		RVR~\cite{zhang2019synthetic} & 0.103 & 28.8 & 21.9 & 0.951 & 0.783 & 0.113 & 28.9 & 22.0 & 0.951 & 0.787 \\
		SteReFo~\cite{busam2019sterefo} &
		0.038 & 32.8 & 26.4 & 0.976 & 0.923 & 0.040 & 32.5 & 26.3 & 0.973 & 0.928 \\
% 		\midrule
		DeepLens~\cite{wang2018deeplens} & 0.068 & 29.5 & 24.9 & 0.945 & 0.897 & 0.055 & 29.8 & 24.9 & 0.957 & 0.911 \\
		DeepFocus~\cite{xiao2018deepfocus} & 0.083 & 33.4 & \underline{30.3} & 0.938 & 0.923 & 0.063 & 35.2 & \underline{31.9} & 0.968 & 0.952 \\
		\midrule
		Ours ($N_{bg}=1$) & \underline{0.011} & \underline{36.7} & 30.0 & \underline{0.989} & \underline{0.951} & \underline{0.019} & \underline{36.8} & 30.5 & \underline{0.986} & \underline{0.956} \\
		Ours ($N_{bg}=3$) & \textbf{0.008} & \textbf{39.3} & \textbf{33.0} & \textbf{0.991} & \textbf{0.963} & \textbf{0.017} & \textbf{38.8} & \textbf{33.0} & \textbf{0.988} & \textbf{0.966} \\
		\bottomrule
	\end{NiceTabular}
	}
	\label{tab:syn_data}
\end{table}

\begin{figure}[t]
    \setlength{\abovecaptionskip}{5pt}
    \scriptsize
	\centering
	\renewcommand\arraystretch{1.0}
	\begin{tabular}{*{6}{c@{\hspace{.7mm}}}}
        \includegraphics[width=0.18\linewidth]{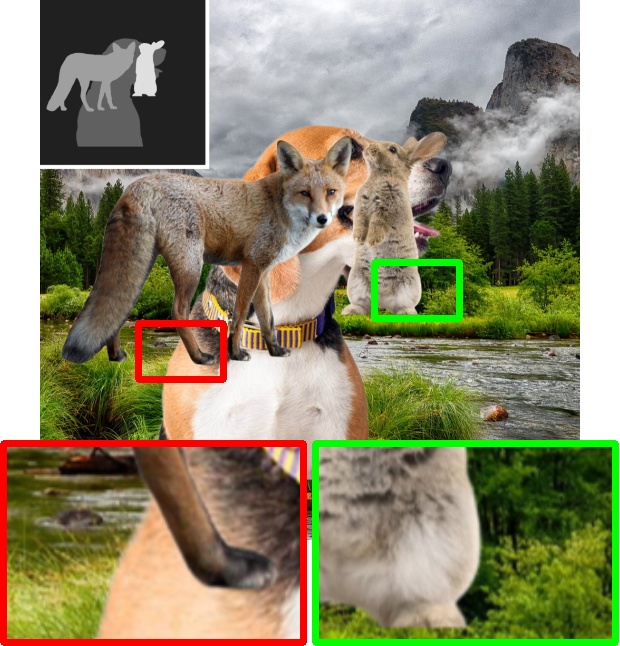} &
        \includegraphics[width=0.18\linewidth]{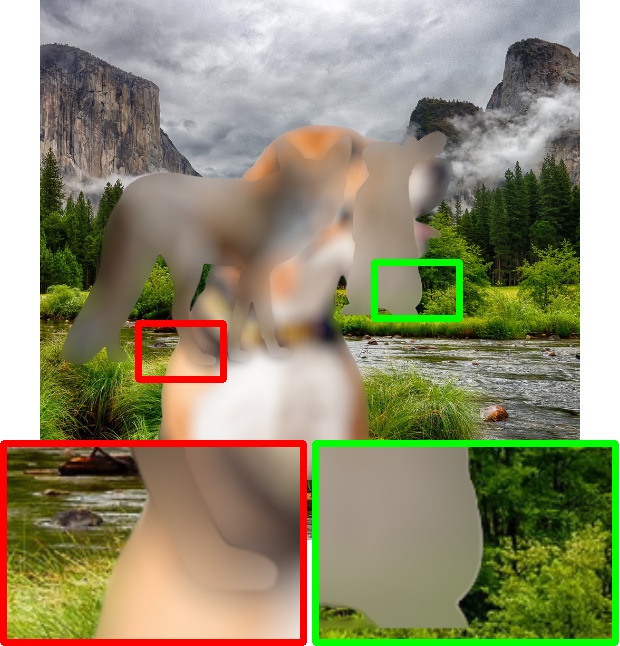} &
        \includegraphics[width=0.18\linewidth]{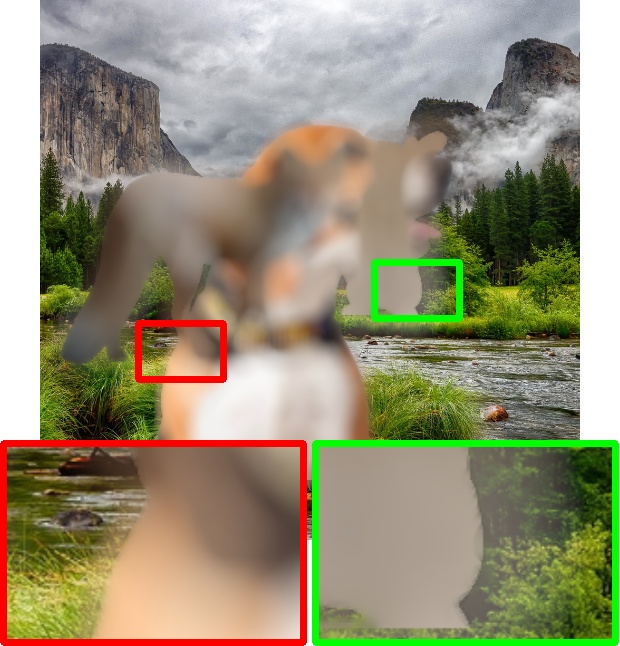} &
        \includegraphics[width=0.18\linewidth]{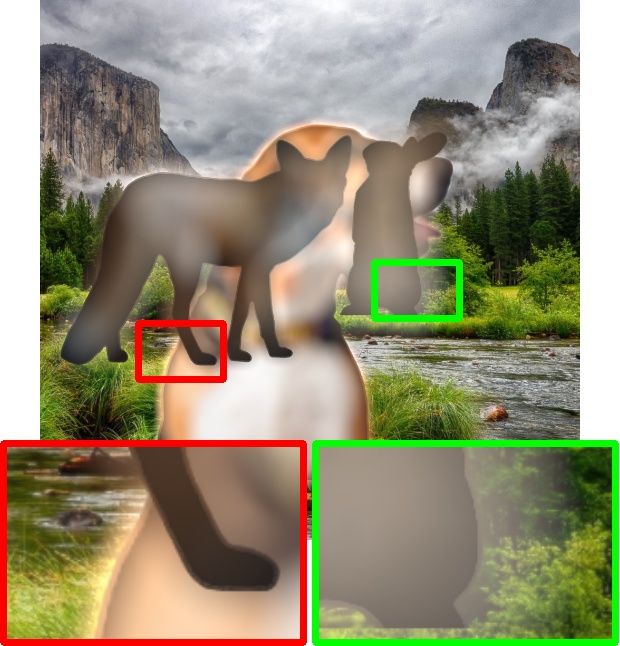} &
        \includegraphics[width=0.18\linewidth]{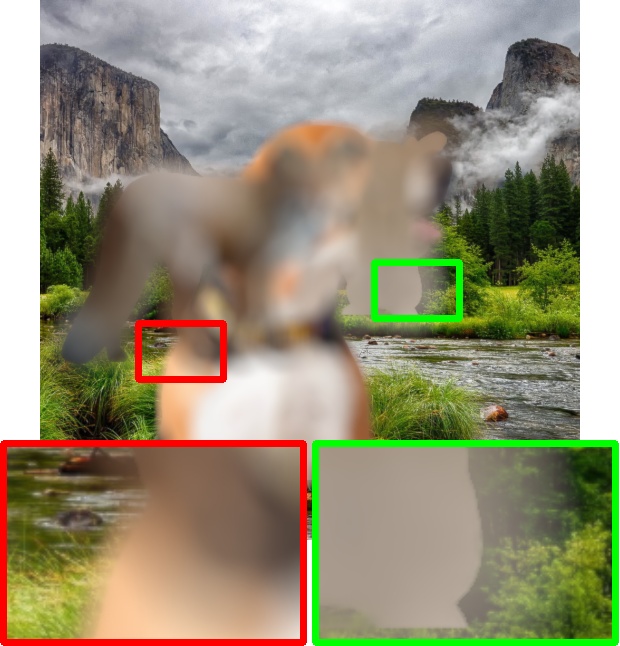} \\
        
        Input & Gather~\cite{wadhwa2018synthetic} & Scatter~\cite{wadhwa2018synthetic} & RVR~\cite{zhang2019synthetic} & SteReFo~\cite{busam2019sterefo} \\
        
        \includegraphics[width=0.18\linewidth]{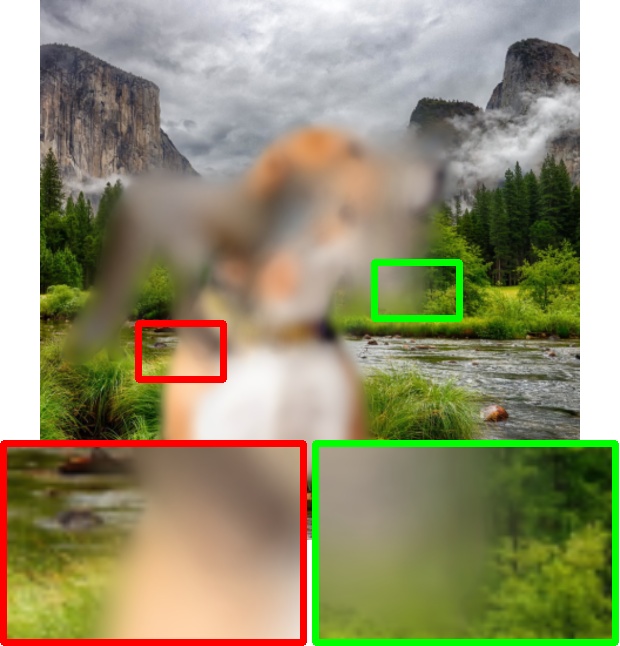} &
        \includegraphics[width=0.18\linewidth]{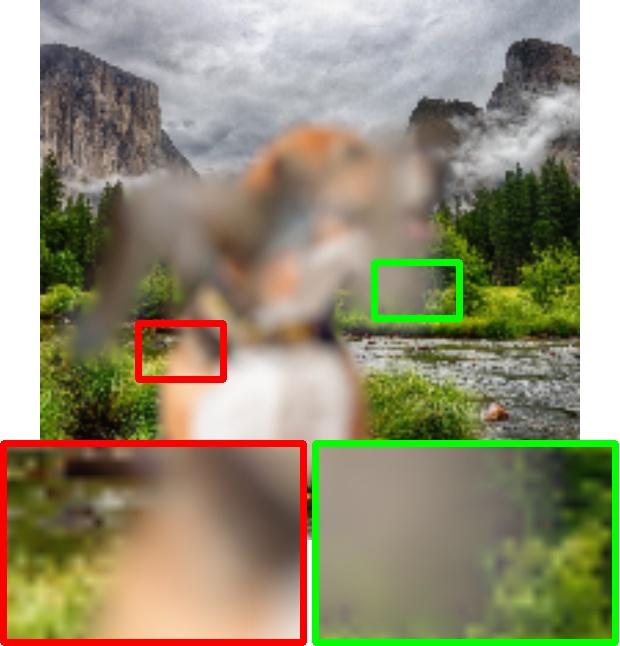} &
        \includegraphics[width=0.18\linewidth]{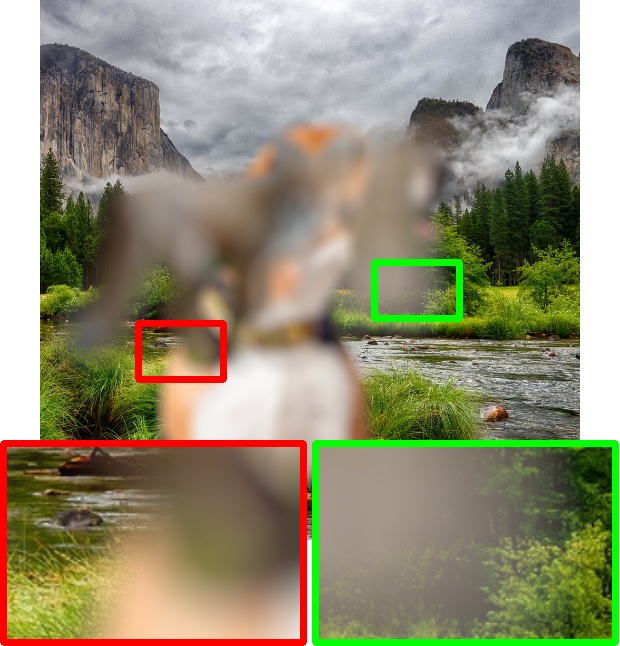} &
        \includegraphics[width=0.18\linewidth]{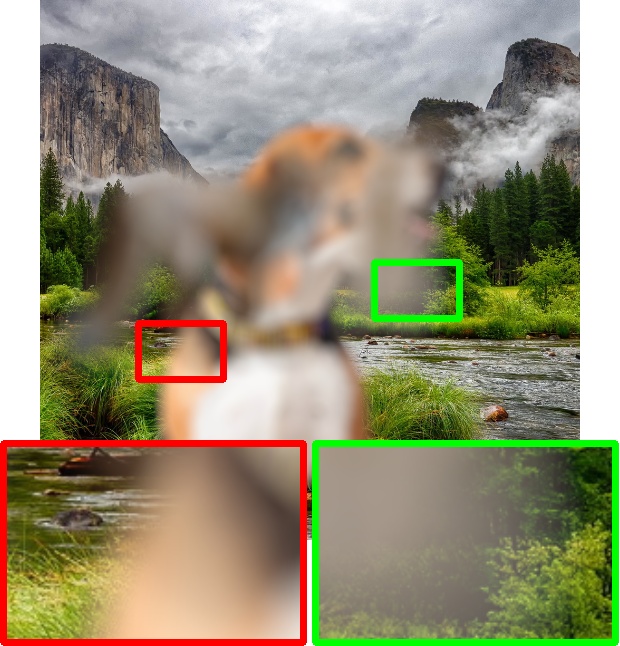} &
        \includegraphics[width=0.18\linewidth]{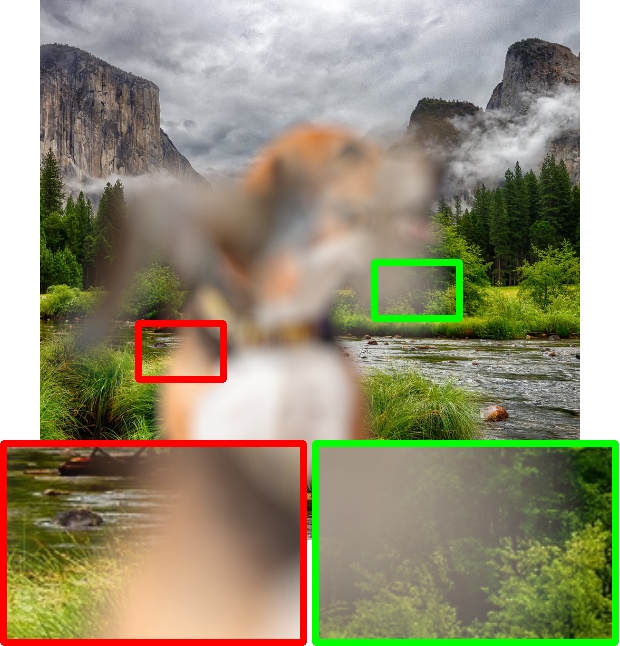} \\
        
        DeepLens~\cite{wang2018deeplens} & DeepFocus~\cite{xiao2018deepfocus} & Ours ($N_{bg}\!=\!1$) & Ours ($N_{bg}\!=\!3$) & GT \\
	\end{tabular}
	\caption{Qualitative results on the synthetic dataset.}
	\label{fig:syn_data}
\end{figure}

\subsubsection{Comparison with State-of-the-Art.} We show quantitative and qualitative results in Table~\ref{tab:syn_data} and Fig.~\ref{fig:syn_data}. One can observe: (\rmnum{1}) Among the physically based rendering methods, pixel-wise methods (Gather~\cite{wadhwa2018synthetic}, Scatter~\cite{wadhwa2018synthetic}) perform well on the overall PSNR and SSIM due to the accurate bokeh rendering in depth-continuous areas, but cause serious color bleeding artifacts at occluding boundaries. For layered methods, SteReFo~\cite{busam2019sterefo} is much better than RVR~\cite{zhang2019synthetic} as SteReFo uses the weight normalization during the rendering. Without this operation, halo artifacts will occur at occluding boundaries, just like RVR. (\rmnum{2}) Neural rendering methods render relatively smooth bokeh effects at boundaries. However, DeepLens~\cite{wang2018deeplens} does not perform very well in metrics. The reason may be that its training data is synthesized by VDSLR~\cite{yang2016virtual}, which applies manual color enhancement and is inconsistent with real rendering rules. DeepFocus~\cite{xiao2018deepfocus} gets high ${\rm PSNR_{ob}}$ and ${\rm SSIM_{ob}}$, but it processes images in low resolution, which leads to poor LPIPS and unpleasant fuzziness and aliasing on refocused plane. (\rmnum{3}) All of the above methods cannot fill the occluded areas with convincing contents, and the rendered results look fake, particularly when the blur amount is large. (\rmnum{4}) Our approach outperforms other methods numerically and creates much more realistic partial occlusion effects. Besides, generating more background images can handle well the continuous occlusion, leading to better performance.

\begin{table}[t]
    % \setlength{\abovecaptionskip}{0pt}
    % \small
	\centering
	\caption{Pairwise comparison results of the user study. A number of more than $50\%$ indicates that our approach gets more votes than the other method.}
	\resizebox{1.0\linewidth}{!}{
    \setlength{\tabcolsep}{5pt}
	\renewcommand\arraystretch{1.0}
	\begin{NiceTabular}{l|cccc|cccc}
		\toprule
		\multicolumn{1}{l}{\multirow{2}{*}[-0.5ex]{Method}} & \multicolumn{4}{c}{Refocusing on the foreground} & \multicolumn{4}{c}{Refocusing on the background} \\
		\cmidrule{2-9}  % \cmidrule(r){2-7} \cmidrule(r){8-13} 
		~ & Scatter\cite{wadhwa2018synthetic} & SteReFo\cite{busam2019sterefo} & DeepLens\cite{wang2018deeplens} & DeepFocus\cite{xiao2018deepfocus} & Scatter\cite{wadhwa2018synthetic} & SteReFo\cite{busam2019sterefo} & DeepLens\cite{wang2018deeplens} & DeepFocus\cite{xiao2018deepfocus} \\
		\midrule
		\midrule
		Ours & \textbf{89.4\%} & \textbf{72.7\%} & \textbf{83.4\%} & \textbf{86.1\%} & \textbf{82.9\%} & \textbf{76.7\%} & \textbf{77.2\%} & \textbf{82.8\%} \\
		\bottomrule
	\end{NiceTabular}
	}
	\label{tab:user_study}
\end{table}

\begin{figure}[t]
    \setlength{\abovecaptionskip}{5pt}
    \scriptsize
	\centering
	\renewcommand\arraystretch{1.2}
	\begin{tabular}{*{6}{c@{\hspace{.3mm}}}}
        
        \includegraphics[width=0.162\linewidth]{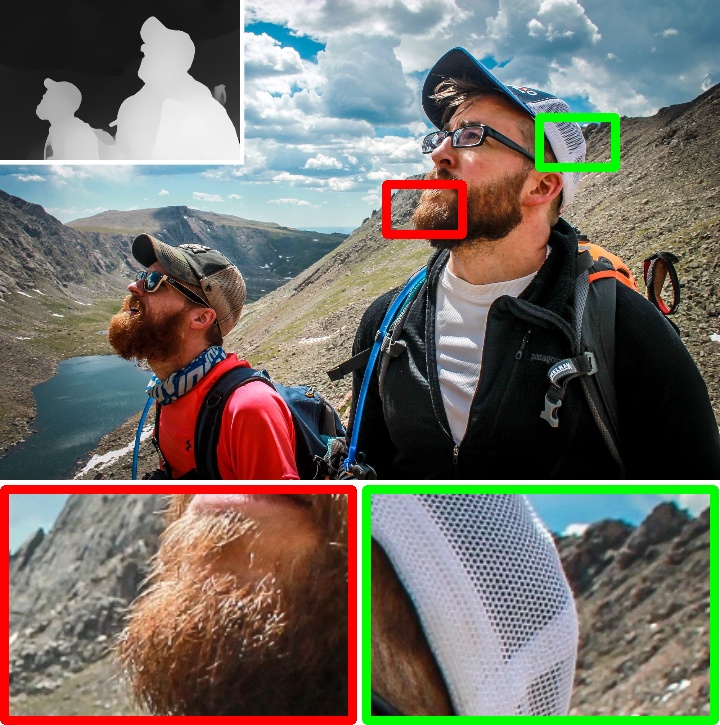} &
        \includegraphics[width=0.162\linewidth]{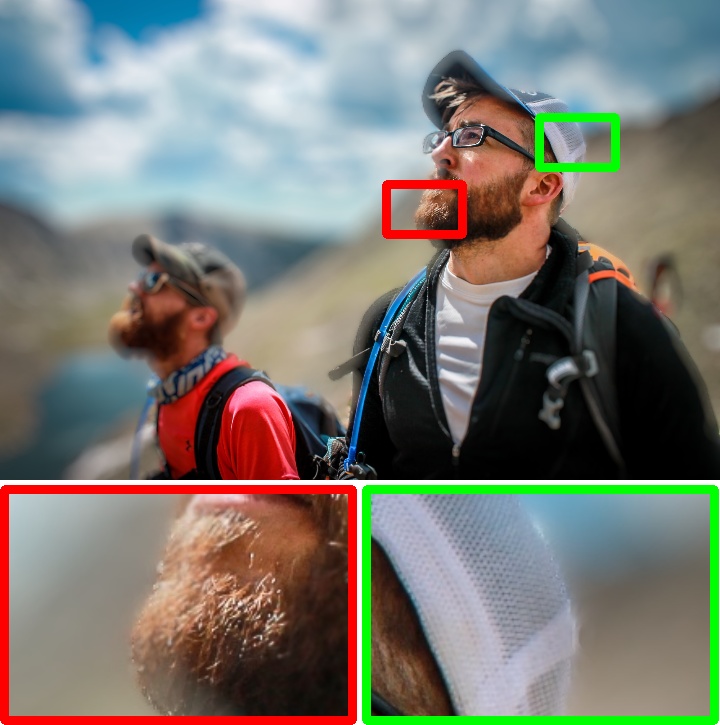} &
        \includegraphics[width=0.162\linewidth]{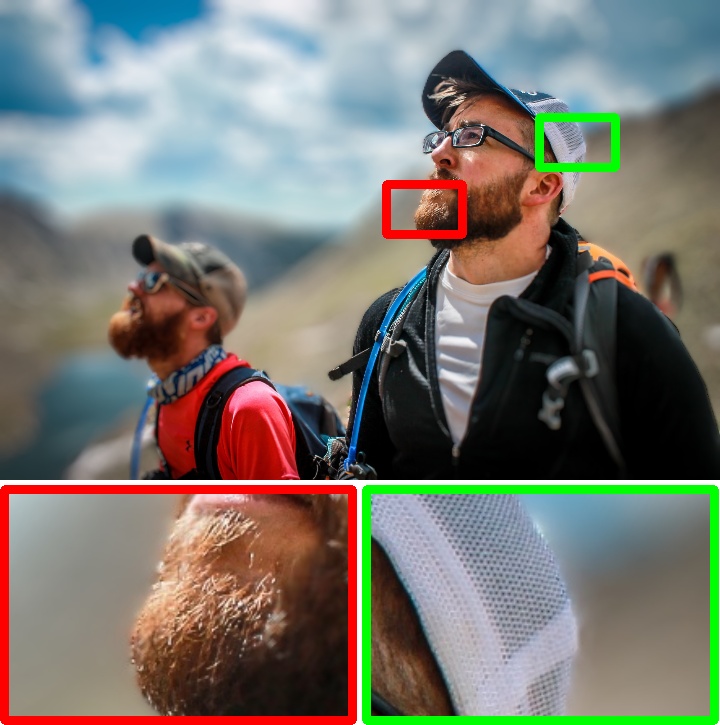} &
        \includegraphics[width=0.162\linewidth]{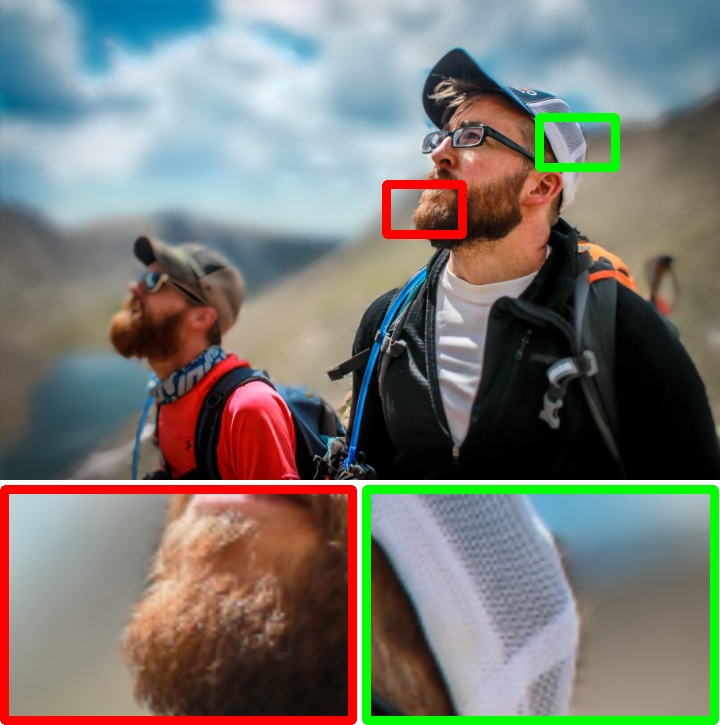} &
        \includegraphics[width=0.162\linewidth]{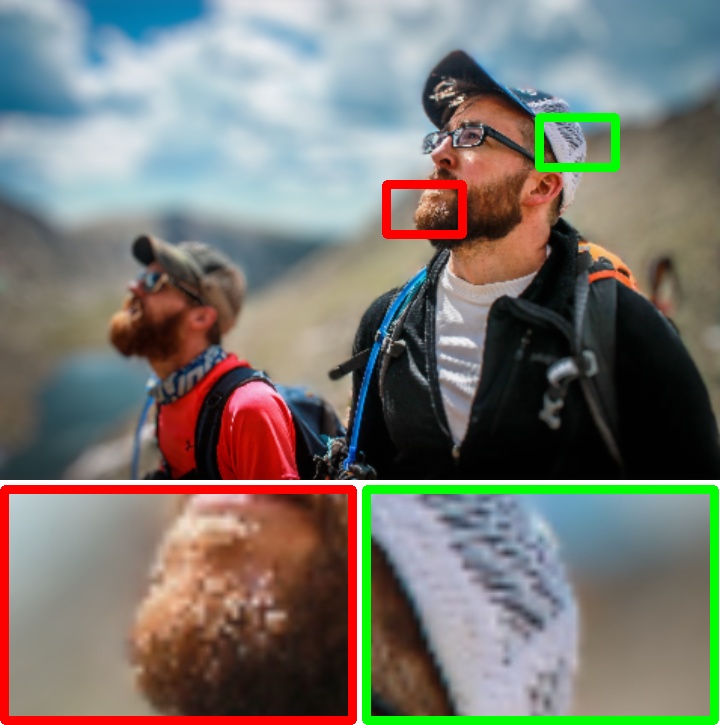} &
        \includegraphics[width=0.162\linewidth]{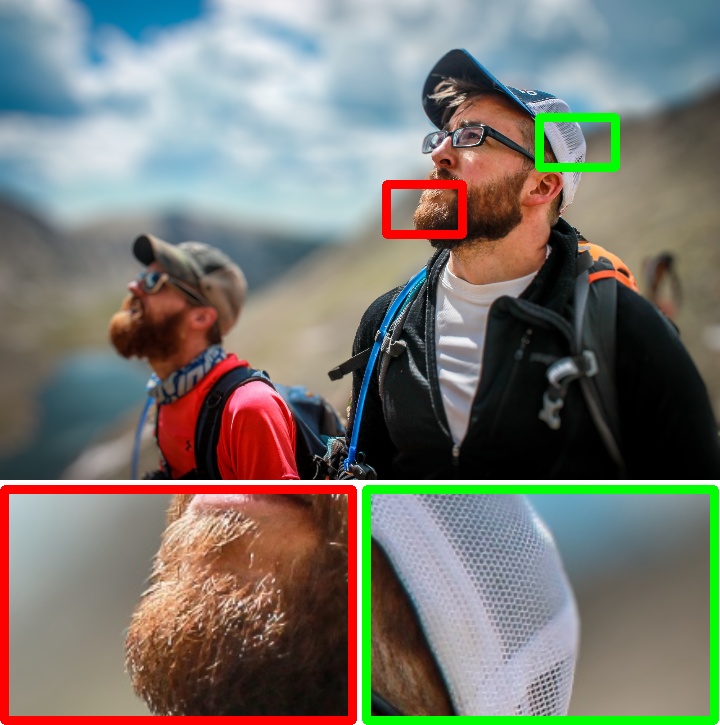} \\
        
        \includegraphics[width=0.162\linewidth]{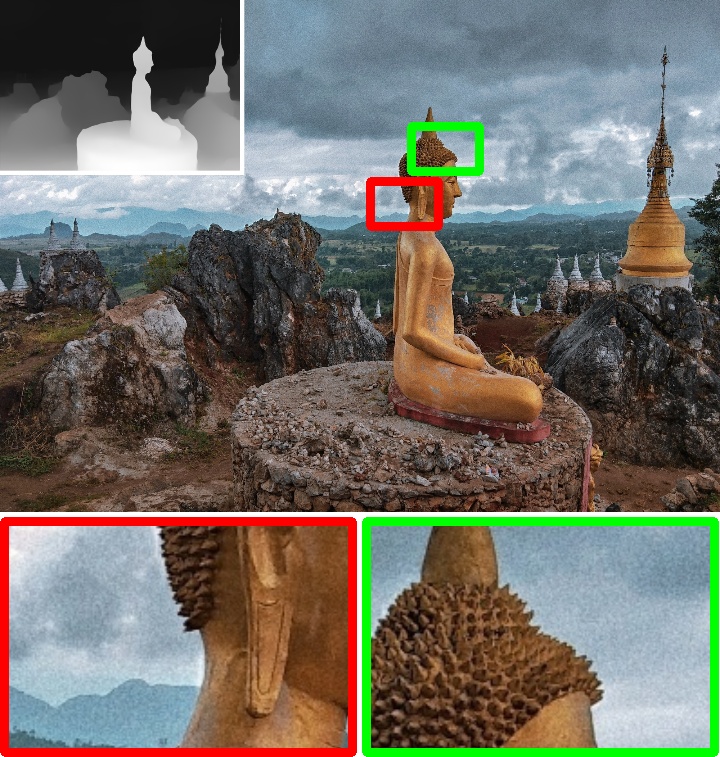} &
        \includegraphics[width=0.162\linewidth]{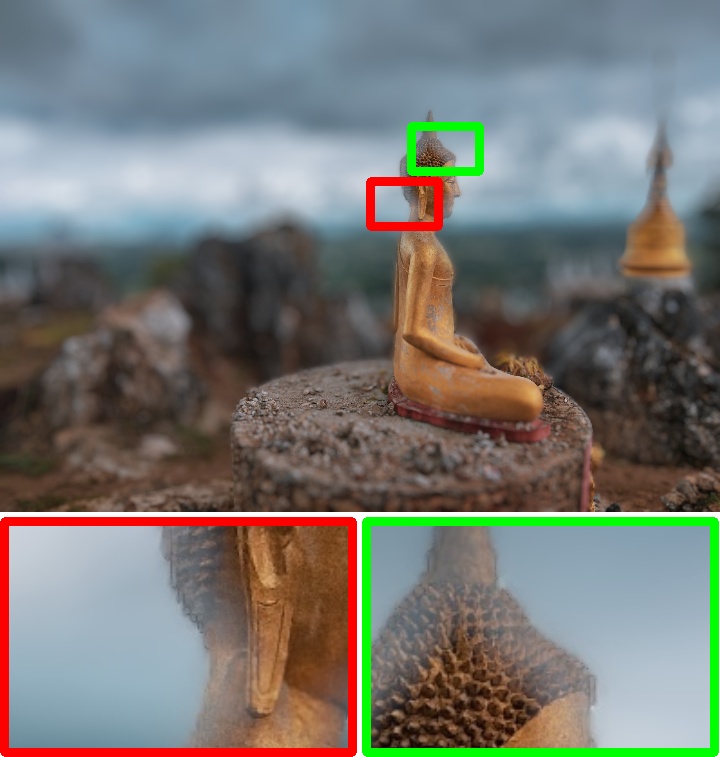} &
        \includegraphics[width=0.162\linewidth]{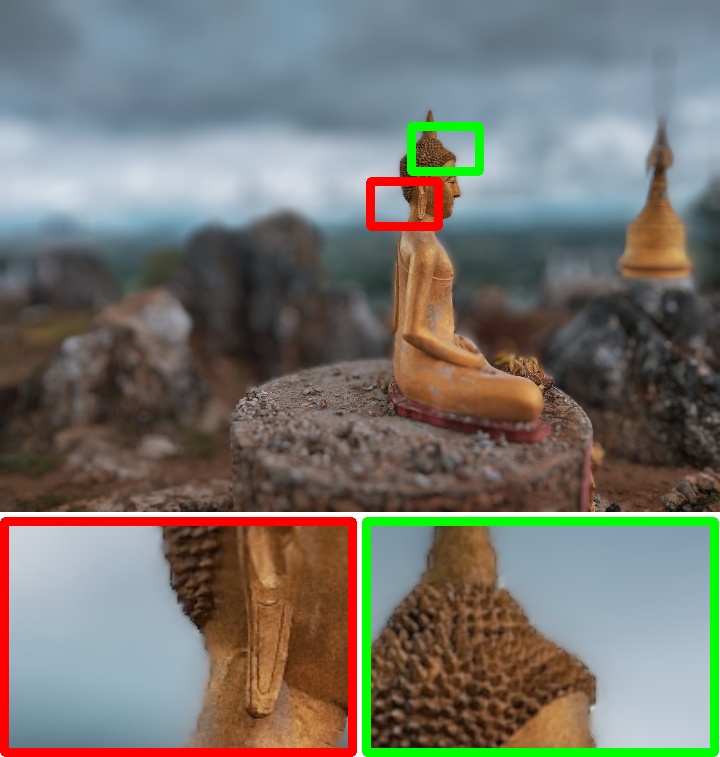} &
        \includegraphics[width=0.162\linewidth]{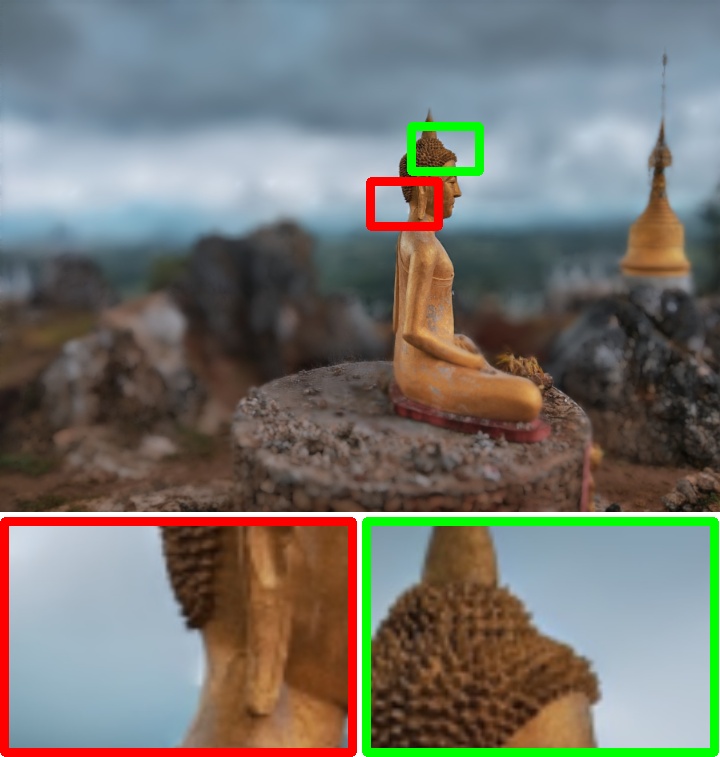} &
        \includegraphics[width=0.162\linewidth]{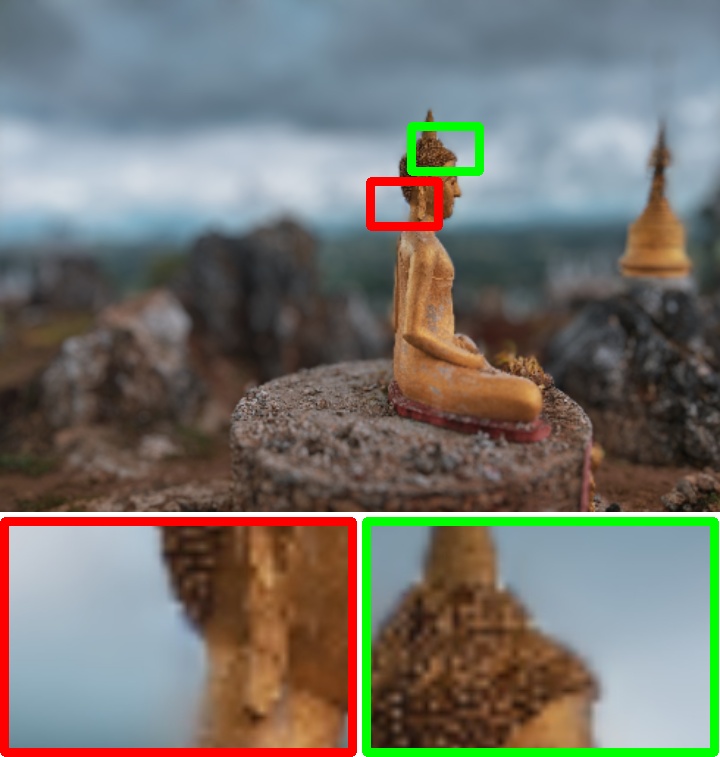} &
        \includegraphics[width=0.162\linewidth]{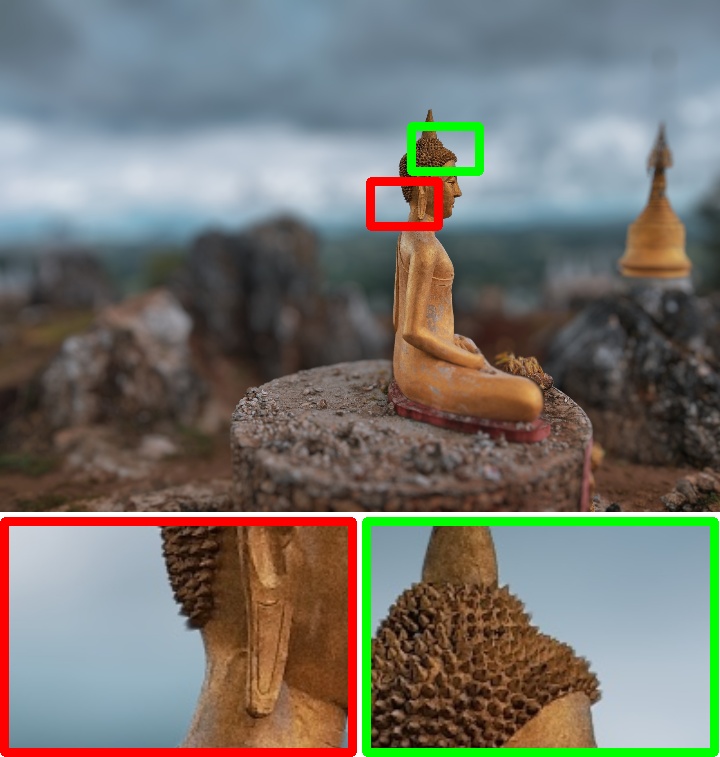} \\
        
        \includegraphics[width=0.162\linewidth]{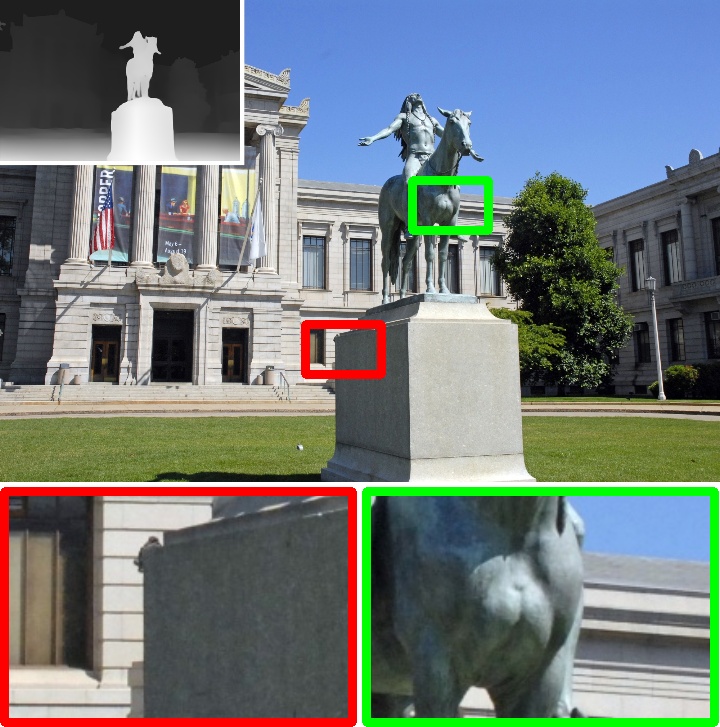} &
        \includegraphics[width=0.162\linewidth]{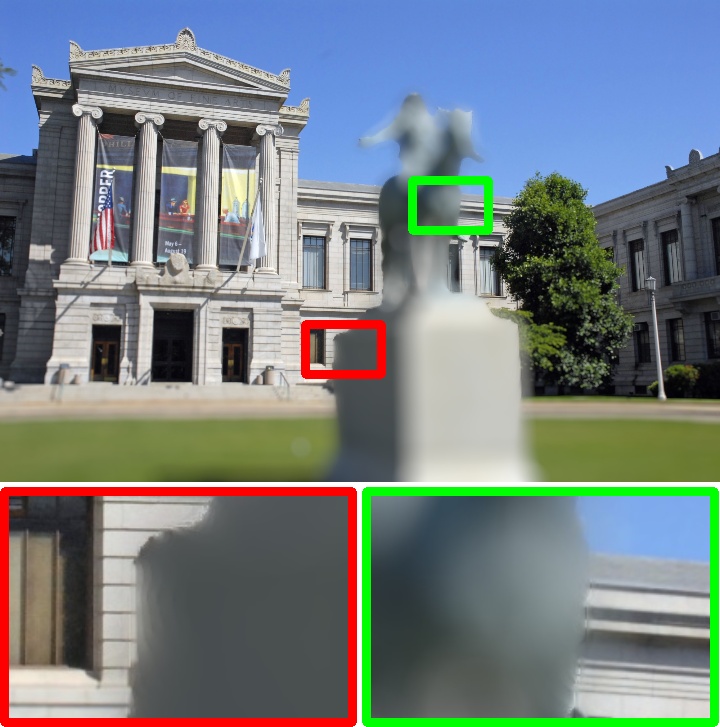} &
        \includegraphics[width=0.162\linewidth]{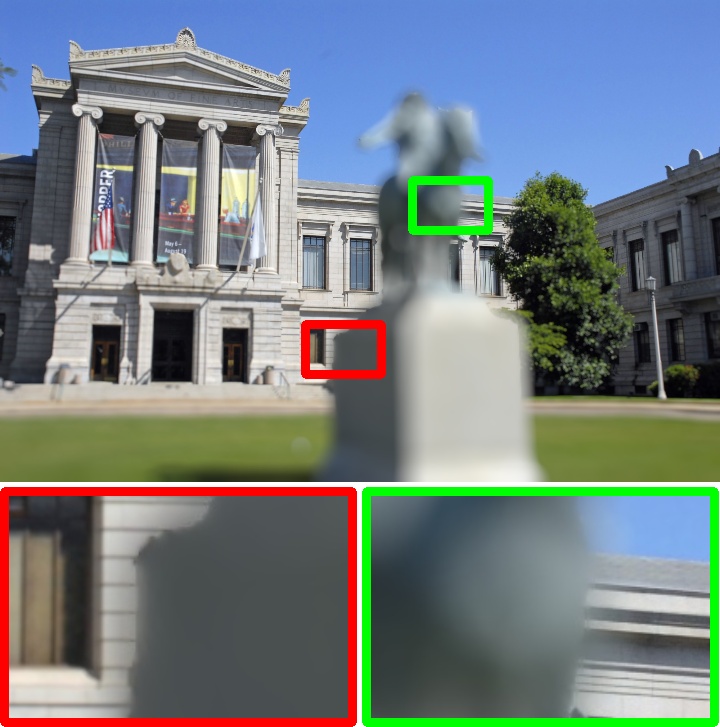} &
        \includegraphics[width=0.162\linewidth]{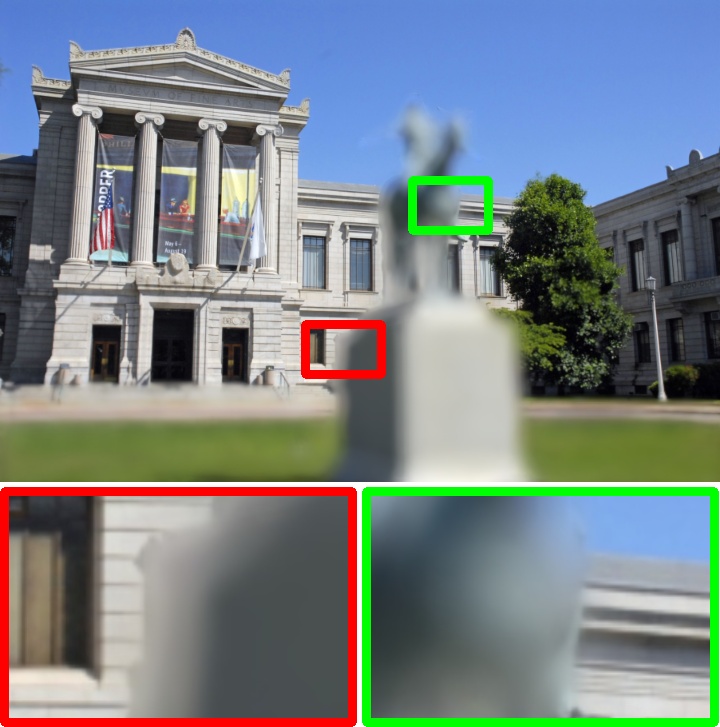} &
        \includegraphics[width=0.162\linewidth]{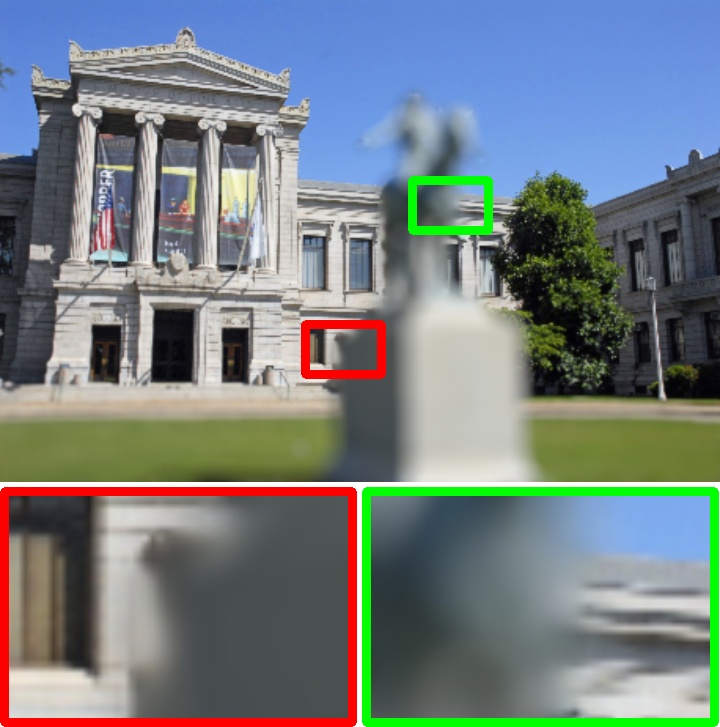} &
        \includegraphics[width=0.162\linewidth]{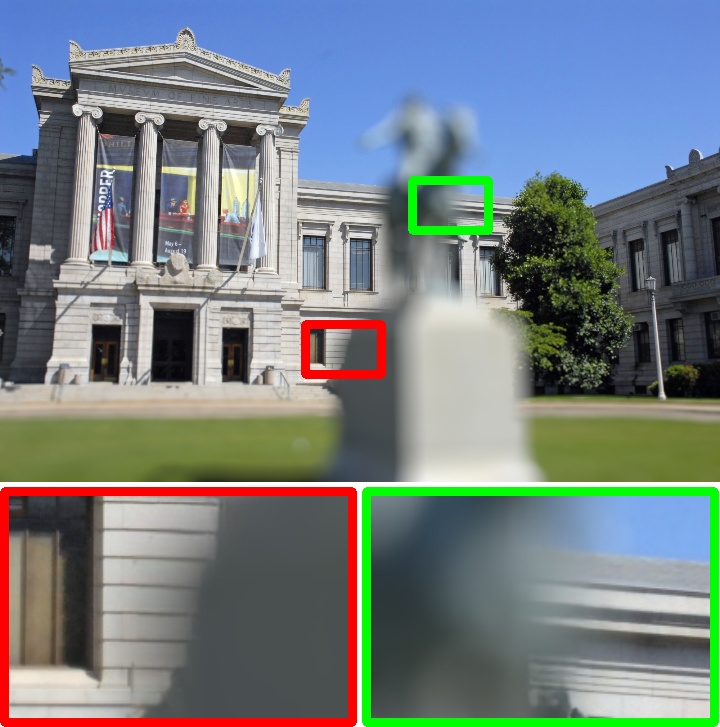} \\

        \includegraphics[width=0.162\linewidth]{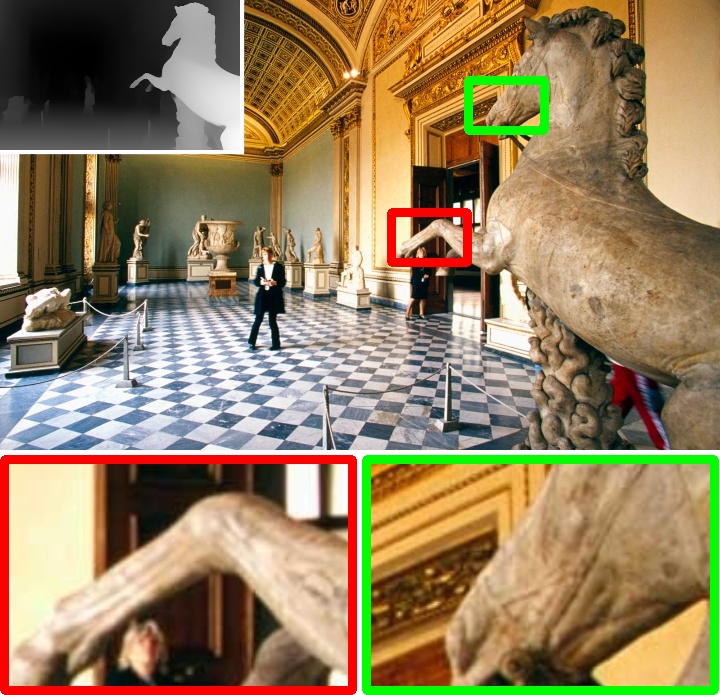} &
        \includegraphics[width=0.162\linewidth]{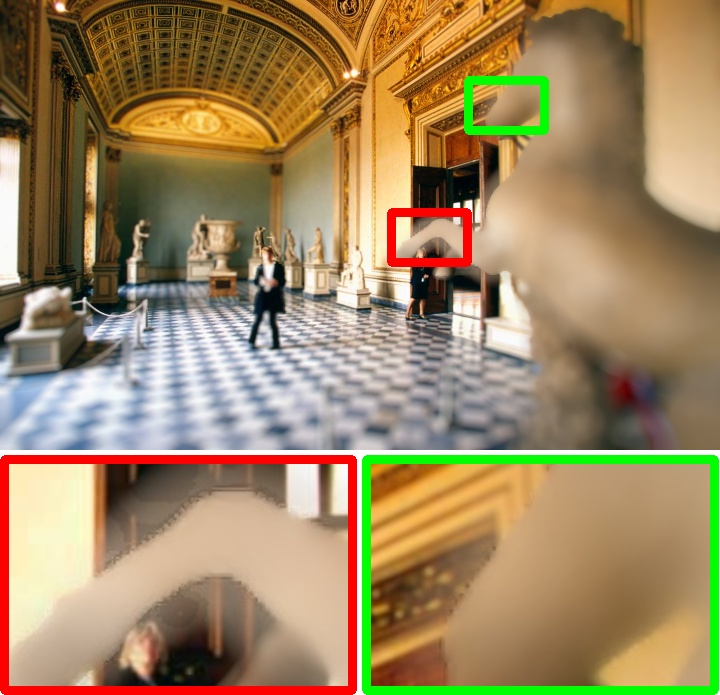} &
        \includegraphics[width=0.162\linewidth]{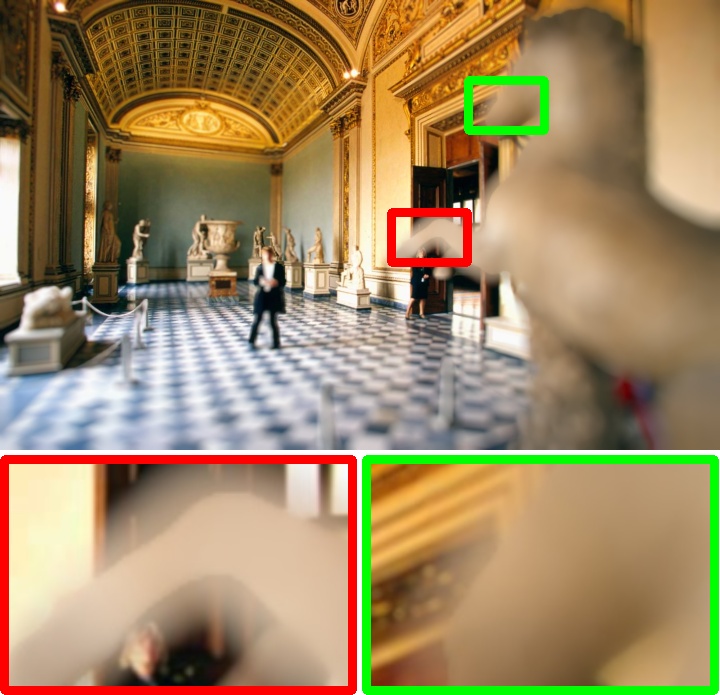} &
        \includegraphics[width=0.162\linewidth]{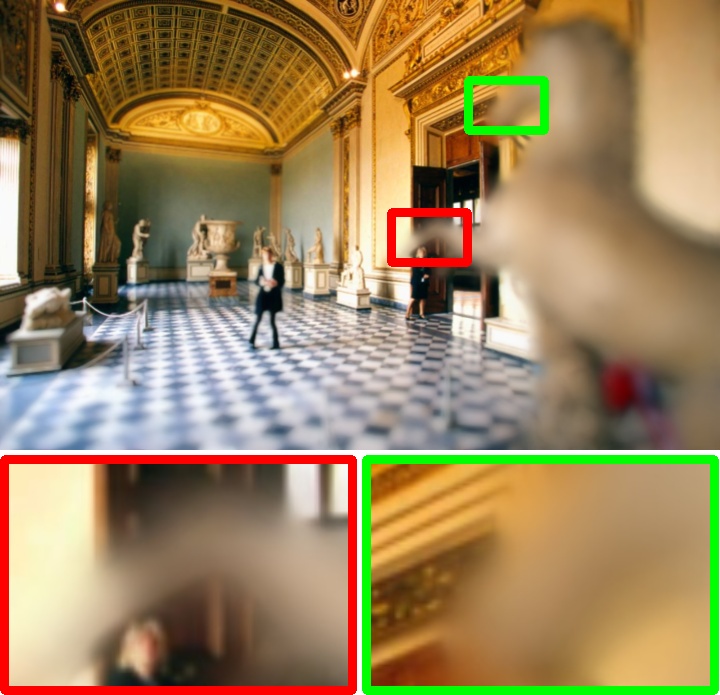} &
        \includegraphics[width=0.162\linewidth]{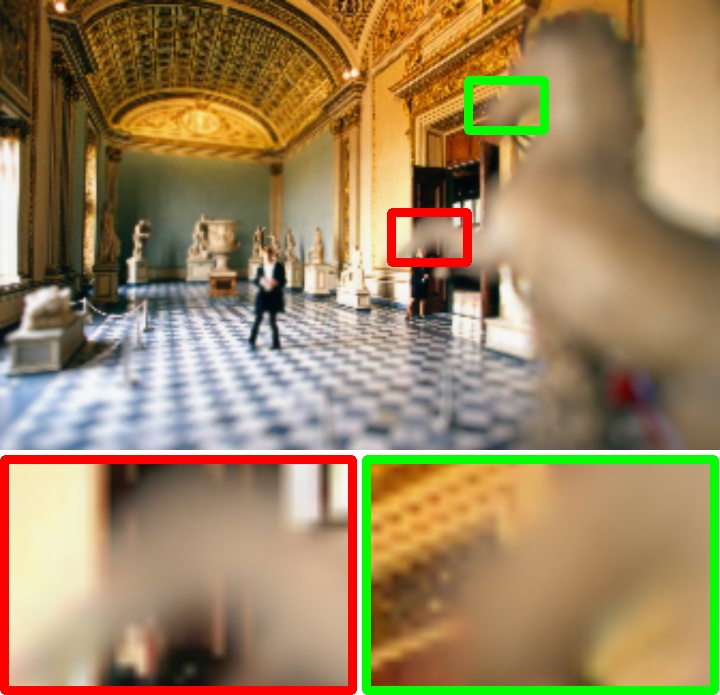} &
        \includegraphics[width=0.162\linewidth]{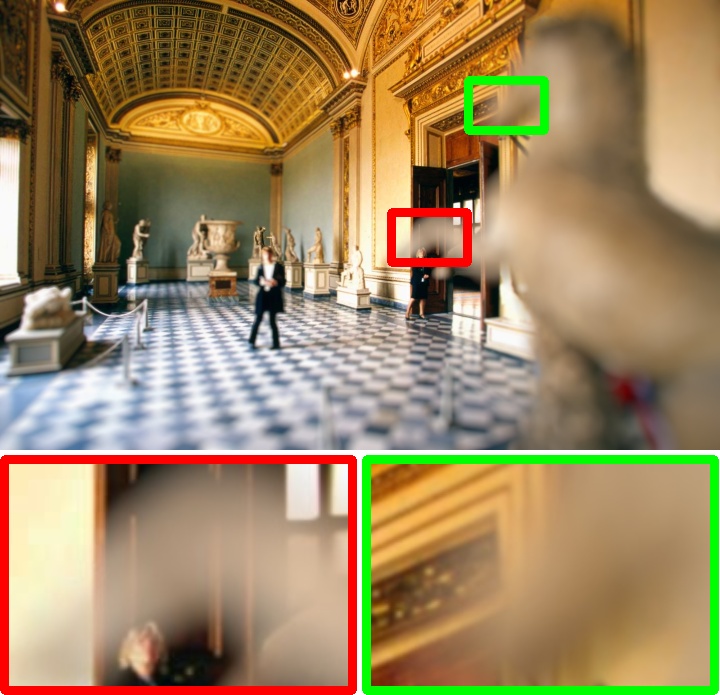} \\
        
        Input & Scatter~\cite{wadhwa2018synthetic} & SteReFo~\cite{busam2019sterefo} & DeepLens~\cite{wang2018deeplens} & DeepFocus~\cite{xiao2018deepfocus} & Ours \\
	\end{tabular}
	\caption{Qualitative results of the user study. The first $2$ rows are refocused on the foreground. The last $2$ rows are refocused on the background.}
	\label{fig:user_study}
\end{figure}

\begin{figure}[t]
    \setlength{\abovecaptionskip}{5pt}
    \scriptsize
	\centering
	\renewcommand\arraystretch{1.2}
	\begin{tabular}{*{8}{c@{\hspace{.5mm}}}}
        
        \includegraphics[width=0.118\linewidth]{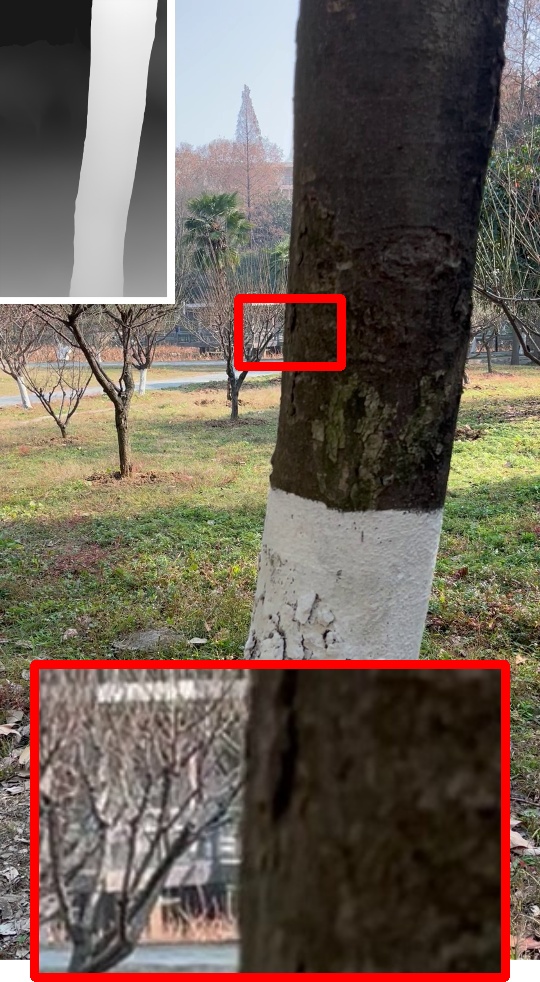} &
        \includegraphics[width=0.118\linewidth]{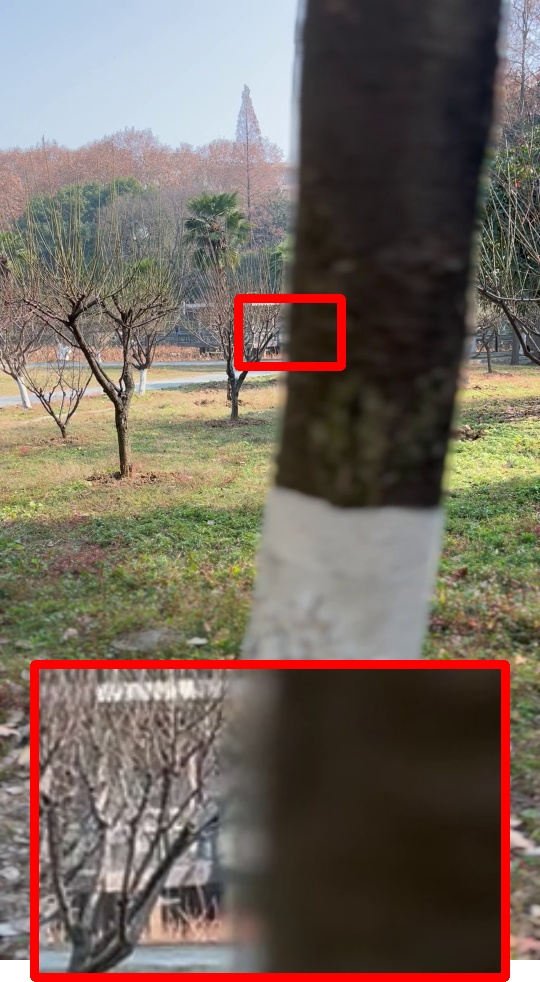} &
        \includegraphics[width=0.118\linewidth]{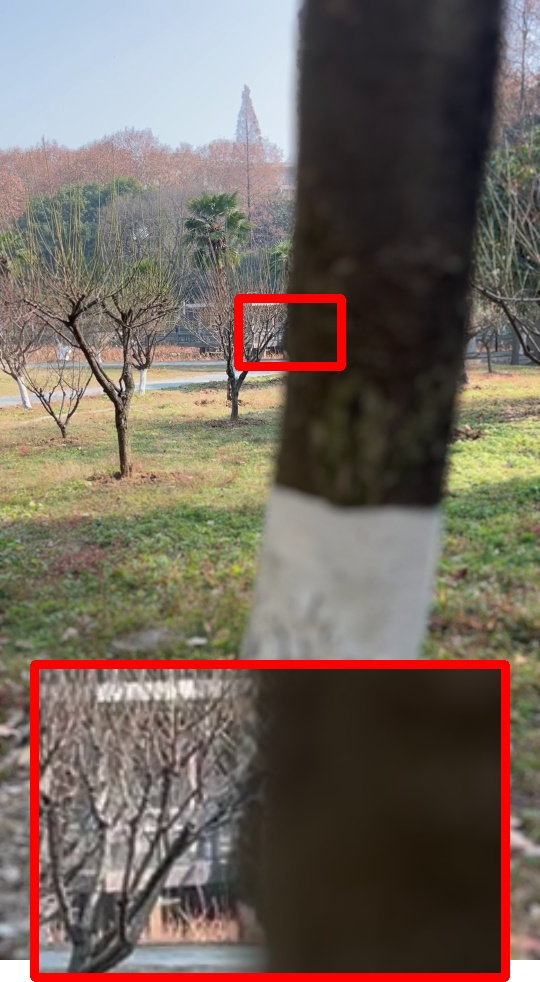} &
        \includegraphics[width=0.118\linewidth]{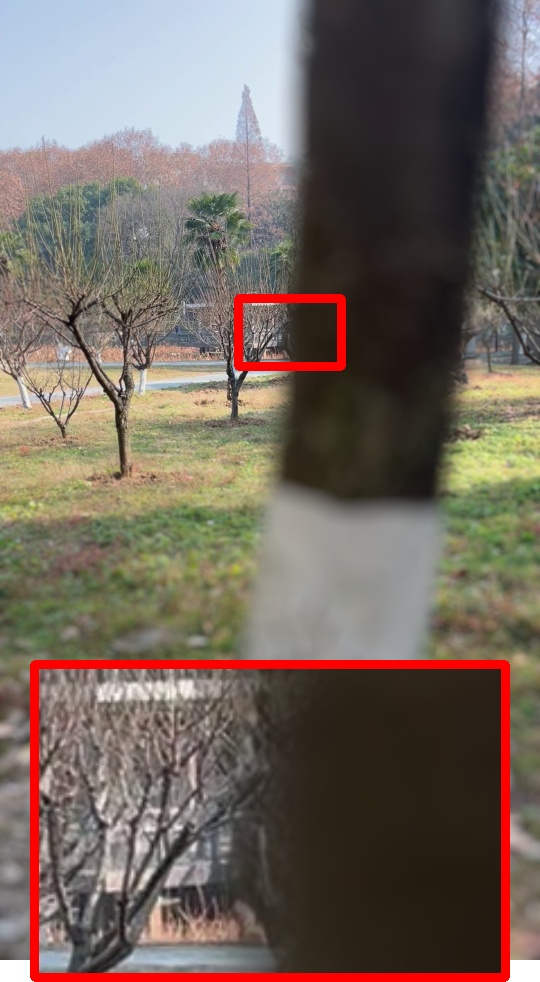} &
        
        \includegraphics[width=0.118\linewidth]{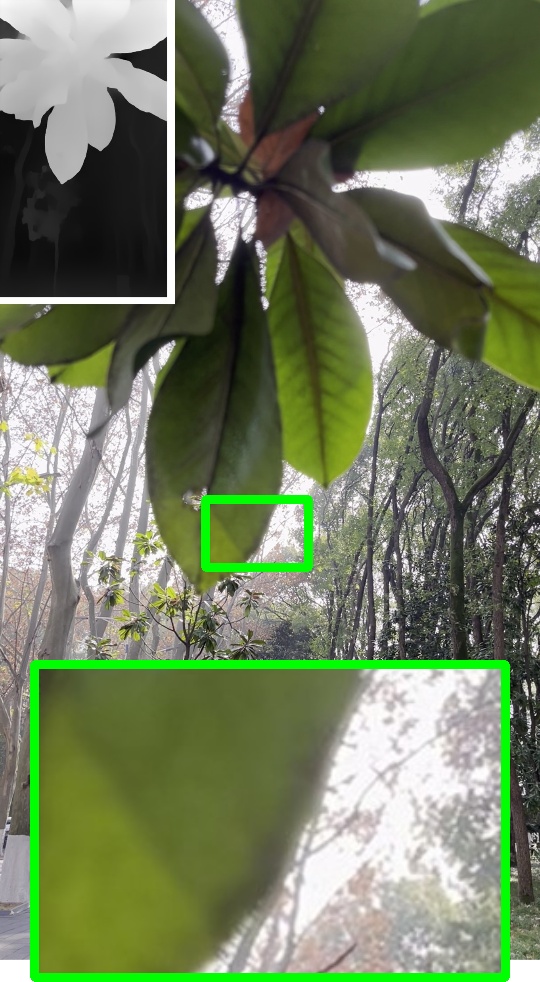} &
        \includegraphics[width=0.118\linewidth]{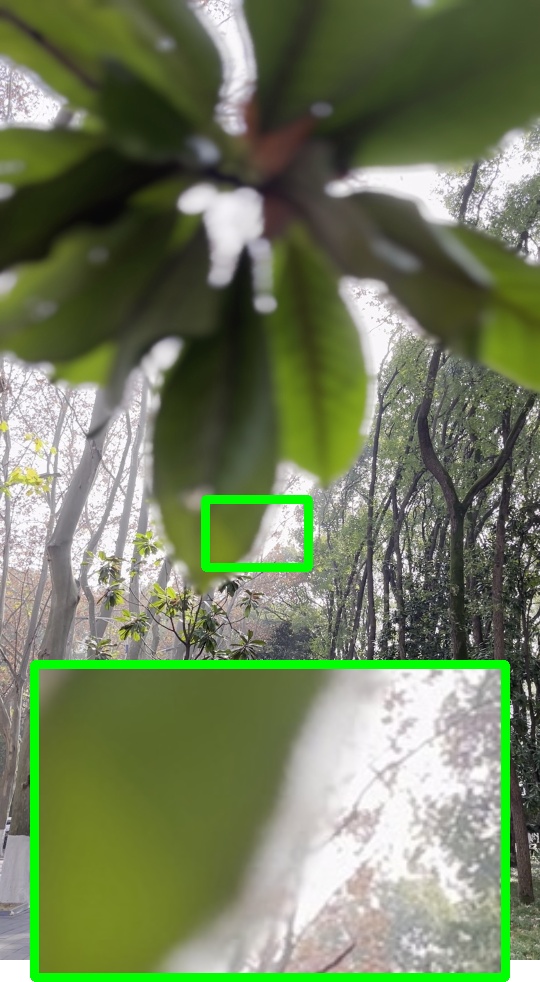} &
        \includegraphics[width=0.118\linewidth]{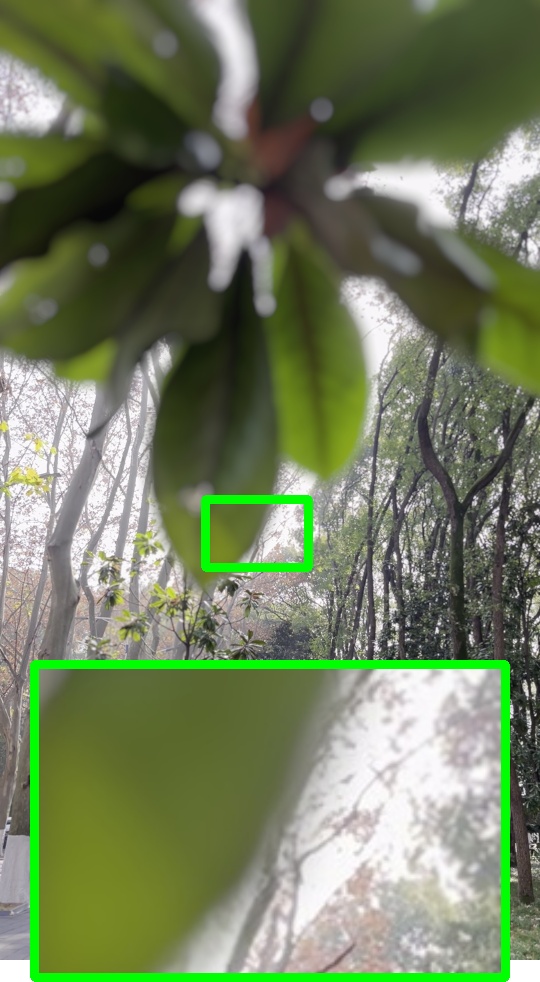} &
        \includegraphics[width=0.118\linewidth]{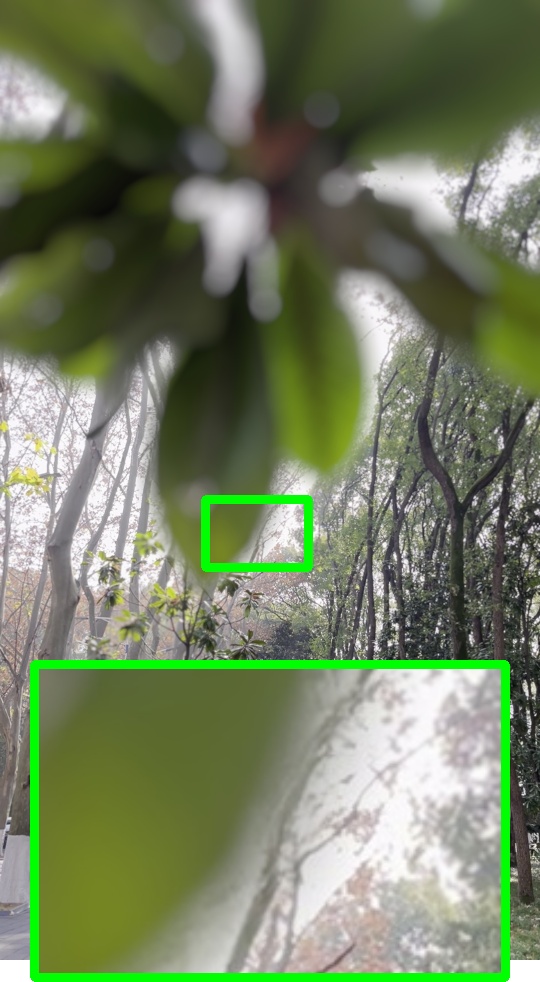} \\
        
        Input & iPhone 13 & Ours & Ours$^{large}$ & Input & iPhone 13 & Ours & Ours$^{large}$ \\
        
	\end{tabular}
	\caption{Comparison with iPhone 13 Cinematic Mode. Image resolution is $1080\times1920$. The superscript $large$ denotes applying the larger blur amount in the rendering.}
	\label{fig:iphone13}
\end{figure}

\subsection{Bokeh Rendering on Real-World Images}
\label{sec:real_world}
In this section, all disparity inputs are predicted by DPT~\cite{ranftl2021vision}, and we use only one inpainted image in the background inpainting module.

\subsubsection{User Study.} Since numerical metrics cannot reflect the perceptual quality of the rendered result, we conduct a user study on real-world images. Specifically, we collect $50$ all-in-focus images from websites, including both indoor and outdoor scenes. The average resolution of images is $1690\times1354$. For each image, we manually select $2$ focal points, one on the foreground and the other on the background. Then, we use different methods to produce bokeh images under the same controlling parameters. During the test, we show each participant $4$ images at a time, including an all-in-focus image, a darker all-in-focus image with the focal point labeled, and two bokeh images in random order produced by our approach and a method randomly selected from Scatter~\cite{wadhwa2018synthetic}, SteReFo~\cite{busam2019sterefo}, DeepLens~\cite{wang2018deeplens} and DeepFocus~\cite{xiao2018deepfocus}. The participant is required to choose the method with more realistic bokeh effects or choose none if it is hard to judge. In addition, each participant needs to complete at least $10$ tests before submitting results. Finally, $99$ people participate in this survey, and there are $2097$ valid pairwise comparisons. The data of the $2$ focal points is counted separately. We show the comparison results in Table~\ref{tab:user_study}, where the number represents the preference of our approach over the other method. One can observe that whether the image is refocused on the foreground or the background, our approach is most favored. Some qualitative results are visualized in Fig.~\ref{fig:user_study}. See the supplementary material for more experimental details and visual results.

It is also worth noting that for other rendering methods, their rankings in this user study differ from the results on the synthetic dataset. For example, Scatter~\cite{wadhwa2018synthetic} gets high metrics on the synthetic dataset, but it is the least preferred in the user study due to the obvious boundary artifacts. This phenomenon demonstrates that the current metrics may fail to accurately measure the perceptual quality of rendered bokeh results. 

\subsubsection{Comparison with iPhone 13 Cinematic Mode.} We further compare MPIB with the latest feature of iPhone 13 - Cinematic Mode, which can freely change the focus point after capturing. From Fig.~\ref{fig:iphone13}, the upper limit of blur amount for iPhone 13 Cinematic Mode is small, and our approach creates more clear and natural bokeh effects in transition areas of the foreground and background.

\begin{table}[t]
    % \small
	\centering
	\caption{Ablation study on the synthetic dataset. %We inpaint $1$ background image in this experiment.
	The best performance is in \textbf{boldface}. Our settings are \underline{underlined}.} 
	\resizebox{1.0\linewidth}{!}{
    \setlength{\tabcolsep}{3pt}
	\renewcommand\arraystretch{1.0}
	\begin{NiceTabular}{l|c|ccccc|ccccc}
		\toprule
		\multicolumn{1}{l}{\multirow{2}{*}[-0.5ex]{Experiment}} & \multicolumn{1}{c}{\multirow{2}{*}[-0.5ex]{Method}} & \multicolumn{5}{c}{Constant disparity for each object} & \multicolumn{5}{c}{Varying disparity for each object} \\
		\cmidrule{3-12}  % \cmidrule(r){3-7} \cmidrule(r){8-12} 
		~ & ~ & LPIPS$\downarrow$ & PSNR$\uparrow$ & ${\rm PSNR_{ob}}\!\!\uparrow$ & SSIM$\uparrow$ & ${\rm SSIM_{ob}}\!\!\uparrow$ & LPIPS$\downarrow$ & PSNR$\uparrow$ & ${\rm PSNR_{ob}}\!\!\uparrow$ & SSIM$\uparrow$ & ${\rm SSIM_{ob}}\!\!\uparrow$ \\
		\midrule
		\midrule
        \multirow{3}{0.17\linewidth}{Inpainting Model} & EC~\cite{nazeri2019edgeconnect} & 0.012 & \textbf{36.8} & \textbf{30.1} & \textbf{0.989} & 0.949 & \textbf{0.019} & \textbf{36.8} & \textbf{30.6} & \textbf{0.986} & 0.955 \\
		~ & MADF~\cite{zhu2021image} & 0.012 & 36.7 & 30.0 & \textbf{0.989} & 0.949 & 0.020 & 36.7 & 30.4 & \textbf{0.986} & 0.955 \\
		~ & \underline{LaMa~\cite{suvorov2022resolution}} & \textbf{0.011} & 36.7 & 30.0 & \textbf{0.989} & \textbf{0.951} & \textbf{0.019} & \textbf{36.8} & 30.5 & \textbf{0.986} & \textbf{0.956} \\
		\midrule
		\multirow{4}{0.17\linewidth}{Number of MPI Planes} & 8 & 0.027 & 34.1 & 28.5 & 0.977 & 0.930 & 0.090 & 31.6 & 27.5 & 0.932 & 0.902 \\
		~ & 16 & 0.015 & 36.0 & 29.6 & 0.986 & 0.946 & 0.044 & 34.8 & 29.5 & 0.969 & 0.939 \\
		~ & \underline{32}  & \textbf{0.011} & 36.7 & 30.0 & \textbf{0.989} & 0.951 & 0.019 & 36.8 & \textbf{30.5} & 0.986 & 0.956 \\
		~ & 64 & \textbf{0.011} & \textbf{36.8} & \textbf{30.1} & \textbf{0.989} & \textbf{0.952} & \textbf{0.018} & \textbf{36.9} & \textbf{30.5} & \textbf{0.989} & \textbf{0.959} \\
        \midrule
		\multirow{2}{0.18\linewidth}{Weight Normalization} & w/o & 0.012 & 36.2 & 29.7 & \textbf{0.989} & 0.950 & \textbf{0.019} & 36.2 & 30.1 & \textbf{0.986} & \textbf{0.956} \\
		~ & \underline{w/} & \textbf{0.011} & \textbf{36.7} & \textbf{30.0} & \textbf{0.989} & \textbf{0.951} & \textbf{0.019} & \textbf{36.8} & \textbf{30.5} & \textbf{0.986} & \textbf{0.956} \\
		\bottomrule
	\end{NiceTabular}
	}
	\label{tab:ablation}
\end{table}

\subsection{Ablation Study} 
To investigate the impact of different components, we present ablation studies on the synthetic dataset. The results are shown in Table~\ref{tab:ablation}. 

\subsubsection{Inpainting Model.} As we use the off-the-shelf inpainting model to generate background image, we compare $3$ inpainting methods: EC~\cite{nazeri2019edgeconnect}, MADF~\cite{zhu2021image}, and LaMa~\cite{suvorov2022resolution}. From Table~\ref{tab:ablation}, one can observe that their performance is similar, showing that our framework can adapt to different inpainting models.

\subsubsection{Number of MPI Planes.} Increasing the number of MPI planes improves the representation ability of scene geometry and reduces the discretization error, but it also brings more runtime consumption at the same time. We finally choose $32$ planes to achieve a trade-off between the performance and the time consumption.

\subsubsection{Weight Normalization.} As discussed in Sec.~\ref{sec:syn_data}, the weight normalization plays an important role in traditional layered rendering methods. However, the performance gain of this operation is not that significant in our layer composition. The reason is that the learned alpha maps cover the invisible background parts and overlap in areas where the disparity changes smoothly, which makes the denominator of Eq.~\ref{eq:bokeh_from_mpi_norm} more inclined to be $1$.

\section{Conclusion}
We have presented an analysis on how to apply the MPI representation and the layer compositing formulation to bokeh rendering, and proposed an MPI-based framework MPIB to render multiple high-resolution bokeh effects and handle realistic partial occlusion effects. 
Despite the fact that this framework works well in general, it still has some limitations, such as the fracture of the connected area due to the plane discretization, and the boundary artifacts caused by terrible inpainting results or the continuous occlusion,
which are shown in the supplementary material. We will address these issues in our future work.

\subsubsection{Acknowledgements.} This work was funded by Adobe.

\clearpage

\bibliographystyle{splncs04}
% \bibliography{BokehRef}
\end{document}